\documentclass[11pt]{article}
\usepackage[top=1.5in,bottom=1.25in,left=1.5in,right=1.5in]{geometry}
\usepackage{graphicx}
\usepackage{amsmath}
\usepackage{bbm}
\usepackage{amssymb}
\usepackage{amsthm}
\usepackage{amsfonts}
\usepackage{wrapfig}
\usepackage{multicol}
\usepackage{float}
\usepackage{pdfpages}
\usepackage[colorlinks=true, urlcolor=blue, linkcolor=blue,citecolor=blue]{hyperref}
\usepackage{natbib}
\usepackage[utf8]{inputenc}
\usepackage{color}

\newtheorem{definition}{Definition}

\theoremstyle{remark}

\theoremstyle{remark}

\newcommand{\omt}[1]{}

\title{The Theory is Predictive, but is it Complete? \\ An Application to Human Perception of Randomness}
\author{Jon Kleinberg\thanks{Cornell University.  Supported in part by a Simons Investigator Award.} \quad \quad Annie Liang\thanks{Microsoft Research and University of Pennsylvania} \quad \quad Sendhil Mullainathan\thanks{Harvard University}}
 
\begin{document}

\maketitle

\begin{abstract}
When we test a theory using data, it is common to focus on
correctness: do the predictions of the theory match what we see in the
data? But we also care about completeness: how much of the predictable
variation in the data is captured by the theory? This question is
difficult to answer, because in general we do not know how much
``predictable variation" there is in the problem. In this paper, we
consider approaches motivated by machine learning algorithms as a
means of constructing a benchmark for the best attainable level of
prediction.

We illustrate our methods on the task of predicting human-generated
random sequences. Relative to an atheoretical machine learning
algorithm benchmark, we find that existing behavioral models explain
roughly 15 percent of the predictable variation in this problem. This
fraction is robust across several variations on the problem. We also
consider a version of this approach for analyzing field data from
domains in which human perception and generation of randomness has
been used as a conceptual framework; these include sequential
decision-making and repeated zero-sum games. In these domains, our
framework for testing the completeness of theories provides a way
of assessing their effectiveness over different contexts; we find that
despite some differences, the existing theories are 
fairly stable across our field domains in 
their performance relative to the benchmark.
Overall, our results
indicate that (i) there is a significant amount of structure in this
problem that existing models have yet to capture and (ii) there are
rich domains in which machine learning may provide a viable approach
to testing completeness.

\end{abstract}

\def\citeasnoun{\cite}

\section{Introduction}

When we test theories, it is common to focus on what one might call 
their {\em predictiveness}: do the predictions of the theory match what we see
in the data?  For example, suppose we have a theory of the labor market
that says that a person's wages depend on 
their knowledge and capabilities. We can test this theory by looking at whether more education indeed predicts higher wages in labor data.
Finding this relationship would provide evidence in support of the theory, but
little guidance towards whether an alternative theory may be even more predictive.
The question of whether more predictive theories might exist\textemdash and how much
more predictive they might be\textemdash points toward a second issue, distinct
from predictiveness, which
we will refer to as {\em completeness}: how close is the performance
of a given theory to the best performance that is achievable in the domain?
In other words, how much of the
predictable variation in the data is captured by the theory?

At a conceptual level, completeness is an important construct
because it  lets us ask 
how much room there is for improving the predictive performance of existing theories. Simultaneously, it helps us to evaluate the predictive performance that has already been achieved. This evaluation is not straightforward, because the same level of predictive accuracy can mean very different things in different problems\textemdash for example,  an accuracy of 55\% is strikingly successful for predicting a (discretized) stock movement based on past returns, but extremely weak for predicting the (discretized) movement of a planet based on the relevant physical measurements.\footnote{For example, the planet's mass, position, velocity, and the masses and
positions of all large nearby bodies.} These two problems differ greatly in the best achievable prediction performance they permit, and so the same quantitative level of predictive accuracy must be interpreted differently in the two domains.

One way to view the contrast between these two problem domains is as follows. In each case, an instance $i$ of the prediction problem 
consists of a vector $x_i$ of measured features, and a hidden label $y_i$ 
that must be predicted.
In the case of astronomical bodies, we believe that the measured
features\textemdash mass, position, velocity, and the corresponding values
for nearby bodies\textemdash are sufficient to make highly accurate 
predictions over short time scales.
In the case of stock prices, the measured features\textemdash
past prices and returns\textemdash are only a small fraction of the 
information that we believe may be relevant to future prices. 
Thus, the variation in stock movements {\em conditioned
on the features we know} is large, while planetary motions are well predicted by known features.

The point then is that predictive error in a theory represents
a composite of two things: first, the opportunity for a better model;
and second, intrinsic noise in the problem due to the limitations
of the feature set.  If we want to understand how much room 
there is for improving the predictive performance of existing theories within a given domain\textemdash
holding constant the set of features that we know how to measure\textemdash
we need a way to separate these two effects.

The challenge is that it is generally very difficult to evaluate the best predictive performance achievable in a given domain.
Are there non-trivial problem domains in which this activity is feasible?
In other words, are there settings that simultaneously (i) contain
complex structure and a rich line of published theories but (ii) are
also tractable enough that we can establish a benchmark of optimal
predictive accuracy for purposes of comparison?

\subsection{A Model Domain: Human Generation of Randomness}
In this paper, we identify such a domain and study theories in it
from the perspective of completeness.
The problem is one with a long history of
research in psychology and behavioral economics: human generation of
random sequences. It is well documented that humans misperceive
randomness \citep{barhillel,LSN},
and this fact is significant not only for its basic psychological interest,
but also for the ways in which misperception of randomness manifests
itself in a variety of contexts: for example, 
investors' judgment of sequences of (random) stock returns \citep{shleifer}, and professional
decision-makers' reluctance to choose the same (correct) option multiple times
in succession \citep{Shue}.

A common experimental framework in this area is to ask human participants
to generate fixed-length strings of $k$ (pseudo-)random coin flips, 
for some small value
of $k$ (e.g. $k = 8$), and then to compare the produced distribution over length-$k$
strings to the output of a Bernoulli process that generates
realizations from $\{H,T\}$ independently and uniformly at random \citep{budescu,distributional}.
We consider the following two natural prediction tasks:
\begin{itemize}
\item {\em Continuation:} We take a string of $k$ pseudo-random coin flips created
by a human participant trying to simulate a Bernoulli process, reveal the
first $k-1$ flips, and ask for a prediction of the $k^{\rm th}$ flip.
\item {\em Classification:} We take a set of $n$ pseudo-random strings
created by a human participant trying to simulate a Bernoulli process,
and $n$ random strings created by a Bernoulli process, and
try to classify each string based on the source (human or Bernoulli)
that produced it.
\end{itemize}
A number of influential existing theories in behavioral economics
provide methods for estimating the probability that different strings are generated
by a human source, and hence lead to predictions for these problems
\citep{Rabin,gambler}.

What is striking is that despite the richness of the underlying
questions, the Continuation and Classification problems are 
behavioral-science questions where the benchmark of optimal
prediction can in fact be feasibly computed.
Optimal predictions for this problem can be made via 
{\em table lookup}, in which we enumerate all $2^k$ strings $s$ 
consisting of 0's and 1's, and for each such string $s$ we record 
the empirical fraction $g(s)$ of human-generated strings in our
sample that are equal to $s$.
With enough samples, this converges to the {\em human distribution}
over the full set of strings.
And from this table of empirical frequencies, it is easy to derive optimal 
predictions for both the Continuation and Classification problems.
For Continuation, this is based on looking at the relative frequency
of $s$ followed by 0 versus $s$ followed by $1$, where $s$ is the 
length-$(k-1)$ prefix we observe;
for Classification, this is based on looking at the human frequency $g(s)$
relative to the Bernoulli probability $2^{-k}$ for a given length-$k$ 
string $s$.

Our analysis in this paper, based on table lookup as a benchmark 
for optimal prediction, thus has a dual motivation.
First, we will uncover a number of new findings about our
substantive domain, the human perception of randomness.
Second, we are able to undertake a case study of theory completeness
for a rich problem,
as discussed at the outset of the paper: given existing theories
and a benchmark for optimal prediction, we can see how close to
optimality the existing theories come, and how this gap varies
for different settings of the question.
We believe that there are a number of domains in the behavioral 
sciences where ``narrow'' feature sets will make this type of baseline
possible. Approaches such as 
\citet{Naecker}, contemporaneous to ours\textemdash which uses machine learning as a benchmark
for behavioral theories of risk and ambiguity\textemdash point to further
potential for this argument. 

\subsection{Overview of the Analysis}
We begin by considering a set of human-generated strings of length 8
over the alphabet $\{H,T\}$ (for ``heads'' and ``tails''),
generated by partipants on Mechnical Turk.
For both the Continuation and Classification prediction problems, we consider
methods that output predictions consisting of probabilities in the
interval $[0,1]$, which are then evaluated relative to the true label of
$0$ or $1$.  We use mean-squared error as our evaluation; thus, 
predicting a probability of 0.5 for all instances would yield
an error of $0.25$.

We find that the existing behavioral models are \emph{predictive}: they 
attain a mean-squared error of $0.249$, which improves (to a statistically significant extent) on the error of 0.250 that we would obtain by random guessing. They are not, however, \emph{complete}. Table lookup attains an error of $0.243$, and relative to this benchmark, the existing models achieve roughly 15\% of the maximum achievable gain over naive guessing for the problem. Thus, there is predictable structure in the problem that has not been fully captured by the existing models. 

We then use this domain to consider
two broad lines of questions related to our notion of theory completeness.
The first is a question of what explains the improvement of table lookup over the behavioral models. In particular, when we say that human-constructed theories
only achieve a relatively small fraction
of the available performance gain over naive guessing, is this 
(a) because they are not using crucial features of the
problem, or (b) because they are not combining them effectively? To separate these possibilities, we take a set of human-constructed features based on research in the area, and apply standard machine algorithms to learn combinations of these features for prediction. We find that in fact these algorithms come close to the performance
of table lookup. Moreover, a substantial amount of the improvement over the behavioral models persists even when these algorithms are restricted to use of only a small number of features (comparable to the number of free parameters in the behavioral models). These results suggest that the answer to the question above may be more (b) than (a)\textemdash the research community approximately knows the ``right'' features for the problem, but may not be combining them as effectively as the machine learning algorithms for the goal of prediction. 

These results bear also on the feasibility of the table lookup
benchmark. While this set of tasks related to the human perception of
randomness made it possible to construct the benchmark of optimal
prediction explicitly, in many domains it will be infeasible in general to
construct a perfect benchmark.  The performance of machine learning
methods such as Lasso regression and decision trees for our task
suggests that in some domains, scalable algorithms
come close to the performance of table lookup, and may serve as
reasonable proxies for the optimal benchmark. (Here too we find
support in the concurrent results of \citeasnoun{Naecker}.)

The second question pertains to the robustness of these results to small variations
in the framing of the task.
In particular, does table lookup succeed by
capturing specific features of the generation of 
length-8 strings of heads/tails 
that do not generalize even to closely related problems?
To address this question, we build a table-lookup predictor 
using the original data of length-8 coin flips, and
then we use this predictor for strings generated 
in a set of related but non-identical domains.
Specifically, we set up prediction problems using binary 
alphabets other than $H$ and $T$, and strings of different 
lengths (using seven flips to predict one additional flip 
at different indices in the string).
We find that in these modified prediction problems, the existing models
produce no more than 20\% of the improvement in prediction error obtained
using table lookup, suggesting that the benchmark and ratio
discovered previously are indeed stable across local problem domains.

\subsection{Applications to Field Data}
Finally, we ask whether our methodology can also be used to
evaluate theory completeness in real-life settings
where human perception of randomness is believed to play a role.
We focus principally on two such settings.

The first is a task involving sequential decision-making\textemdash
specifically, data on baseball umpires calling balls and strikes.
\citet{Shue} find that umpire calls are negatively auto-correlated: in aggregate, umpires tend to avoid long runs
of the same call (i.e. calling many strikes in a row or many balls in a row).
Within this setting, we ask: knowing only an umpire's most
recent $k-1$ calls, how well can we predict the current call?

Our second field study uses data from repeated play of Rock-Paper-Scissors on the Facebook app Roshambull, which was collected by \citet{RPS}. Each unit of observation is a game, where a game consists of a sequence of matches that conclude when one of the players wins two matches. In this setting, we ask: knowing only the choices (rock, paper, or scissors) that a player made in his or her
first $k-1$ matches, how well can we predict the choice in the current match? 

In both problems, we find that table lookup can achieve significant
gains over naive guessing.
Moreover, when we evaluate the completeness of the model based on \citet{gambler}, we find its completeness in both domains to be qualitatively
similar to what we obtained in our basic experimental framework.
This shows that the completeness of the model is relatively stable across domains that are quite different, and in all
cases there is significant room for theories to achieve stronger
predictive gains.

Taken together, our results suggest that (1) there is a significant
amount of structure in the problem of predicting human generation of
randomness that existing models have yet to capture, and 
(2) our approach via the optimal predictive benchmark allows for
evaluation of the completeness of theories in the given domain. 
Such an approach can be
applied more generally in settings where this benchmark can be
feasibly determined or approximated.

\section{Primary Testbed: Human Generation of Fair Coin Flips} \label{sec:main}
\subsection{Data} \label{sec:data}
We use the platform Mechanical Turk to collect a large dataset of
human-generated strings designed to simulate the output of a
{\em Bernoulli(0.5) process}, in which each symbol in the string
is generated from $\{H,T\}$ independently and uniformly at random.
Our main experiment includes 537
subjects, each of whom produced 50 binary strings of length eight,
attempting to generate them
as if these strings were the realizations of 50 experiments in
which a fair coin was flipped eight times. In a second experiment, an
additional 101 subjects were asked to generate 25 binary strings
of length eight. The task was described to subjects using the text
below:

\begin{quote}We are researchers interested in how well humans can produce randomness.  A coin flip, as you know, is about as random as it gets.  Your job is to mimic a coin. We will ask you to generate 8 flips of a coin.  You are to simply give us a sequence of Heads (H) and Tails (T) just like what we would get if we flipped a coin.\\

\noindent Important: We are interested in how people do at this task. So it is important to us that you not actually flip a coin or use some other randomizing device.
\end{quote}

\noindent To discourage use of an external randomizing device, we gave subjects 30 seconds to generate each string. To incentive effort, we told subjects that payment would be approved only if their strings could not be identified as human-generated with high confidence.\footnote{Subjects were informed: ``To encourage effort in this task, we have developed an algorithm (based on previous Mechanical Turkers) that detects human-generated coin flips from computer-generated coin flips. You are approved for payment only if our computer is not able to identify your flips as human-generated with high confidence."} The complete set of directions can be found in Appendix \ref{instructions}. 

Despite these incentives, some subjects did not attempt to mimic a random process, generating for example the same string in each of the fifty rounds. In response to this, we removed all subjects who repeated any string in more than five rounds.\footnote{This cutoff was selected by looking at how often each subject generated any given string, and finding the average ``highest frequency" across subjects. This turned out to be 10\% of the strings, or five strings. Thus, our selection criteria removes all subjects whose highest frequency was above average.} This selection eliminated 167 subjects and 7,400 strings, leaving a final dataset with 471 subjects and 21,975 strings. We check that our main results are not too sensitive to this selection criteria, considering two alternative choices in Appendix \ref{appDiffCut}\textemdash first, keeping only the initial 25 strings generated by all subjects, and then, removing the subjects whose strings are ``most different" from a Bernoulli process under a $\chi^2$-test. The results remain quantitatively the same.

Throughout, we identify Heads with `1' and Tails with `0,' so that each string is an object in $\{1,0\}^8$. The 21,975 strings are aggregated into a single dataset, which induces an empirical distribution over $\{1,0\}^8$. This observed human distribution over strings turns out to be statistically different from a true Bernoulli(0.5) process: we can reject the hypothesis that the data is generated from a uniform distribution over $\{1,0\}^8$ under a $\chi^2$-test with $p\approx 0$.

\begin{figure}[h]
\begin{center}
\includegraphics[scale=.35]{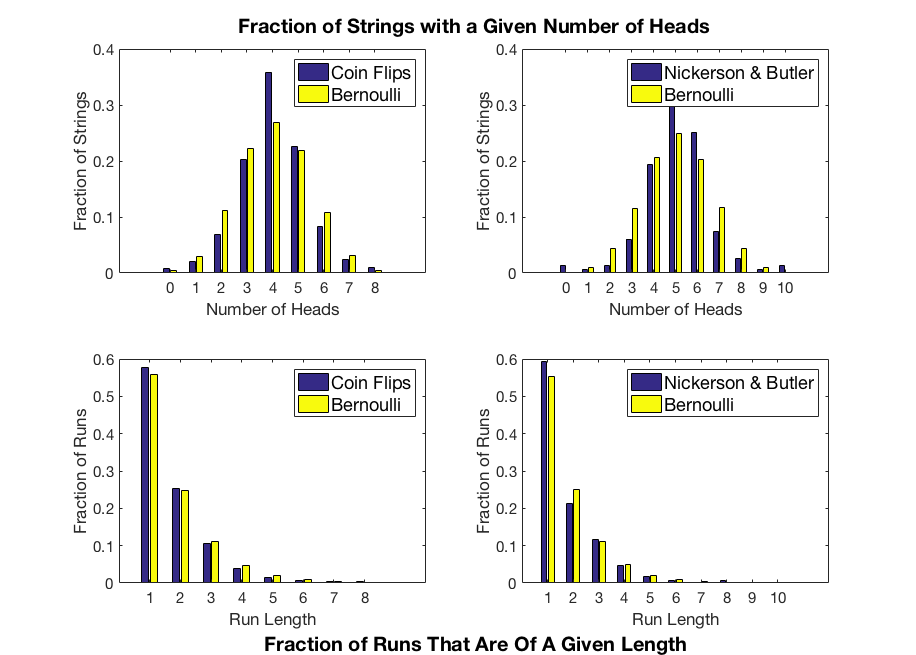}
\end{center}
\caption{\footnotesize{(a) Top row: the fraction of generated strings that include $m$ Heads, where $m$ is the label on the $x$-axis. \textit{Left}\textemdash comparison of MTurk data (purple) with simulated Bernoulli strings (yellow); \textit{Right}\textemdash comparison of Nickerson \& Butler (2009) data (purple) with simulated Bernoulli strings (yellow). (b) Bottom row: the fraction of runs that are of length $m$, where $m$ is the label on the $x$-axis. \textit{Left}\textemdash comparison of MTurk data (purple) with simulated Bernoulli strings (yellow); \textit{Right}\textemdash comparison of Nickerson \& Butler (2009) data (purple) with simulated Bernoulli strings (yellow).}}
\label{fig:heads}
\end{figure}

Moreover, the ways in which the observed distribution differs from a
Bernoulli process are consistent with the literature
\citep{budescu,distributional}. For example, subjects exhibit an
over-tendency to alternate (52.68\% of flips are different from the
previous flip, as compared to an expected 50\% in a Bernoulli(0.5)
process), an under-tendency to generate strings with ``extreme" ratios
of Heads to Tails (see the top row of Figure \ref{fig:heads}), and an
under-tendency to generate strings with long runs (see the bottom row
of Figure \ref{fig:heads}). Furthermore, subjects exhibit strong
context-dependency: the probability of reversal depends not only on the
immediately previous flip, but on the pattern of several prior. Table
\ref{tab:three} lists the frequencies with which each possible
three-flip pattern is followed by `1,' where the strings are sorted
according to this quantity in decreasing order.
These
frequencies are compared with the data in \citet{budescu} (using Table
1 from \citet{gambler}).

\begin{table}[h]%
	\caption{The empirical probability of Heads conditional on the three previous flips.}
	\label{tab:three}
	\bigskip
	\begin{minipage}{\columnwidth}
\begin{center}
\begin{tabular}{cccccc}
\hline
&&&Our data & \citet{budescu}& Bernoulli\\
\hline

         0   &1 &         0  &  0.5995 & 0.588& 0.5\\
                  1 	&         0     &    0 &   0.5406 & 0.62& 0.5 \\
         0      &   0 &  1 & 0.5189 & 0.513& 0.5\\

         0      &   0 &         0 &   0.5185& 0.70 & 0.5 \\
         \hline
             1    &1&    1   & 0.4811 & 0.30& 0.5 \\
         0  &  1   & 1 &    0.4595& 0.38& 0.5\\

        1    &1   &      0   & 0.4528 & 0.487& 0.5\\
        1   &       0&    1 &    0.4415 & 0.412& 0.5 \\
        \hline
\end{tabular}
\end{center}
		\centering
	\end{minipage}
\end{table}%

The difference between the probabilities with which `$000$' and
`$111$' are followed by `1' is significantly smaller in our data than
in \citet{budescu}; besides this, however, we find that these
conditional probabilities are similar. Notably, the strings that are more likely to be followed by `1' in our data are also more likely to be followed by `1'  in the \citet{budescu} data (compare the top four rows and bottom four rows in Table \ref{tab:three}). These similarities suggest that the
dataset we collected is typical of its kind.

\subsection{Existing Models} \label{sec:behavioral-models} Several
frameworks have been proposed for modeling human misperception of
randomness, and we will consider two influential approaches proposed
in \citet{Rabin} and \citet{gambler}. Although both of these
frameworks are models of mistaken \emph{inference} from data, and not
directly models of human generation of random sequences, they are
easily adapted to our setting, as we will discuss
below.\footnote{These adaptations are consistent with comments made in
the original papers, in the context of relating these models to the
empirical literature discussed in Section \ref{sec:data}.}

In \citet{Rabin}, subjects in the underlying model
observe independent, identically distributed (i.i.d.) signals, 
but mistakenly believe the signals to be
negatively autocorrelated. Specifically, subjects observe a sequence
of i.i.d.\ draws from a Bernoulli($\theta$) distribution, where
$\theta \in [0,1]$ is an unknown rate drawn from distribution $\pi$.
Although subjects know the correct distribution $\pi$ over the
Bernoulli parameter $\theta$, they have a mistaken belief about the
way in which the realized rate $\theta$ determines the signal process.
Subjects believe that the observed signals are instead drawn
\emph{without replacement} from an urn containing $\theta N$ `1'
signals and $(1-\theta)N$ `0' signals, so that a signal of `1' is less
likely following observation of `0', and vice versa. For tractability, it is additionally assumed that the urn is ``refreshed" every other round,
meaning that the composition is returned to its original composition of $\theta N$ `1' signals and
$(1-\theta)N$ `0' signals.

To use this model in our setting, we modify it in two ways: first, since subjects are informed that the coin they should mimic is fair, we fix the prior distribution $\pi$ over rates so that subjects believe $\theta = 0.5$ with certainty; second, we relax the assumption that the urn is refreshed deterministically every other round, adding a second parameter $p \in [0,1]$, which determines the probability that the urn is refreshed. Thus, in the revised model, subjects generate random sequences by drawing without replacement from an urn that is initially composed of $0.5N$ `1' balls and $0.5N$ `0' balls, and is subsequently refreshed with probability $p$ before every draw.

\citet{gambler} introduce a second framework for modeling human misperception of randomness. The following simple version of their model can be applied to predicting generation of random sequences: each subject generates flip $s_1$ from a Bernoulli(0.5) distribution, and then each subsequent flip $s_k$ according to  
\[s_k \sim \mbox{Ber}\left(0.5 - \alpha \sum_{t=0}^{k-2} \delta^t (2 \cdot s_{k-t-1}-1)\right),\]
where the parameter $\delta \in \mathbb{R}_+$ captures a (decaying) influence of past flips, and the parameter $\alpha\in \mathbb{R}_+$ measures the strength of negative autocorrelation.\footnote{We make a small modification on the \citeasnoun{gambler} model, allowing $\alpha, \delta \in \mathbb{R}_+$ instead of $\alpha, \delta \in [0,1)$.} Notice that past realizations of `1' reduce the probability that the $k$-th flip is `1', and past realizations of `0' increase this probability. Thus, like the previous model, \citet{gambler} predicts generation of negatively autocorrelated sequences.

\subsection{Prediction Tasks}  \label{sec:predict-task}
We test these theories by looking at how well they predict the dataset of human-generated strings described in Section \ref{sec:data}. We consider two tests. In the first test, which we call \emph{Continuation}, we ask how well we can predict a subject's eighth flip given the first seven flips. A prediction rule for this problem is any function
\begin{equation} \label{eq:cont}
f: \{0,1\}^7 \rightarrow [0,1]
\end{equation}
that map the initial seven flips into a probability that the next flip is `1'. Given a test dataset $\{s^i\}_{i=1}^n$ of $n$ strings, we evaluate the error of the prediction rule $f$ using:
\begin{equation*}
\label{predacc}
\frac1n \sum_{i=1}^n \left( s^i_8 - f(s_{1:7}^i)\right)^2.
\end{equation*}
where $s_8$ is the eighth flip in string $s$, and $f(s_{1:7})$ is the predicted probability that the eighth flip is `1' given initial sequence $s_{1:7}$. This loss function, \emph{mean-squared error}, penalizes (quadratic) distance from the best prediction. Notice that if subjects are truly generating strings from an i.i.d.\ Bernoulli(0.5) distribution, then no prediction rule can improve in expectation upon a prediction error of 0.25.

In the second test, which we call \emph{Classification}, we seek to separate strings generated by human subjects from strings generated by a Bernoulli(0.5) process. A prediction rule in this problem is any map 
\begin{equation} \label{eq:class}
c: \{0,1\}^8 \rightarrow [0,1]
\end{equation}
from strings of length eight into a probability that the string was generated by a human subject. Given a test dataset $\{s^i\}_{i=1}^n$ of $n$ strings, we evaluate error by producing an equal number of Bernoulli strings, and finding
\begin{equation*}
\label{classacc} 
\frac{1}{2n} \sum_{i=1}^{2n} \left( c^i - c \left(s^i\right)\right)^2,
\end{equation*}
where $c^i=1$ if the true source of generation for string $s^i$ was a human subject, and $c^i=0$ otherwise. As above, if the human-generated strings are consistent with a Bernoulli(0.5) process, then we cannot improve on an expected prediction error of 0.25. 

As a brief remark, we note that these tests are different from one
another: a model of human generation can perform well on one
prediction problem and poorly on the other. This is because
Continuation asks how well we can predict the probability that a
string ends in `1,' \emph{conditional} on the first seven entries
in the string, while Classification asks how well we can predict the
{\em unconditional} distribution over entire strings. 
For example, if the human distribution over strings were to contain
non-Bernoulli structure in its first few entries, but were
essentially uniform over the final entry independently of what
had preceded it, then it would be possible to perform well
in Classification but not in Continuation.

If we assume that strings are generated according to either of the models described in Section \ref{sec:behavioral-models}, then there is a ``best" prediction rule that minimizes expected prediction error (see Appendix \ref{appendix:predrule} for more detail).  We can therefore test these models by seeing how well their corresponding prediction rules perform in Continuation and Classification. Specifically, we estimate the free parameters of these models on training data and report their out-of-sample prediction errors (tenfold cross-validated) in Table \ref{tab:results}.\footnote{We randomly partition the data into ten equally-sized subsets, estimate the free parameters of the model on nine subsets (the training set), and predict the strings in the tenth (the test set). The reported prediction error is an average over the ten possible choices of the test set (from the ten folds), and the reported standard error is the standard deviation of the prediction errors across the test sets. This is a common approximation to the standard error for a cross-validated loss.}

Throughout, we compare these errors with a naive baseline that corresponds to random guessing\textemdash that is, we predict that the next flip is `1' with probability 0.5 for all initial substrings in the Continuation task, and we classify each string as human-generated with probability 0.5 in the Classification task. We find that the \citet{Rabin} and \citet{gambler} models are predictive:  their prediction errors are between $0.2491$ and $0.2495$, all of which improve on the error of 0.250 that we would obtain by random guessing. This improvement is statistically significant for the model based on \citet{gambler}.

\begin{table}[H]%
	\caption{\citet{Rabin} and \citet{gambler} are predictive: they improve upon the prediction error achieved by guessing at random.}
	\label{tab:results}
	\bigskip
	\begin{minipage}{\columnwidth}

\begin{center}
\begin{tabular}{ccc}	
\hline
	& Continuation & Classification\\
	\hline
	Naive  & 0.25 & 0.25 \vspace{2mm}\\
Rabin (2002) &0.2494 &     0.2493\\
&(0.0007)& (0.0008)\vspace{2mm}\\
Rabin and Vayanos (2010) & 0.2492& 0.2488\\
&(0.0007)&(0.0006)\vspace{1mm}\\
\hline
\end{tabular} 
\end{center}
		\centering
	\end{minipage}
\end{table}%

But the \emph{margin} of improvement over guessing at random is very small (no larger than 0.0012), and the gap between the best prediction errors and a perfect zero is large. This raises the question of how significant these improvements are.

In order to answer this, we need a benchmark against which to evaluate the improvement. Crucially, this benchmark should not be perfect prediction:  deviations from a true i.i.d. process make it possible to improve upon the naive baseline of 0.25, but the observed process is far from deterministic. Conditioning on initial flips alone, there is a limit to how well we can hope to predict in these problems, so a more suitable benchmark is the \emph{best possible} prediction error.

\subsection{Benchmark} \label{sec:benchmark}
Our proposed approach for constructing such a benchmark for this problem is to use a \emph{table lookup} algorithm, in which we enumerate all $2^k$ binary strings and record 
the empirical frequency of each string.  Given enough samples, this table of empirical frequencies approximates the ``human distribution"
over the full set of strings.   And from this table, we can derive optimal 
predictions for both the Continuation and Classification problems.\footnote{The table lookup prediction error is a consistent estimator for the \emph{irreducible error} in the problem, also known as the \emph{Bayes error rate}.} 

\begin{definition}[Table Lookup]  
\label{TableLookup}
Let $g$ be the empirical distribution over strings in the training data. The table lookup continuation rule is
\begin{equation}
\label{TL}
f_{TL}(s) = \frac{g(s1)}{g(s)}
\quad \forall \,\, s \in \{1,0\}^7. 
\end{equation}
where `$s1$' is the concatenation of the string $s$ and the outcome `1'. The table lookup classification rule is 
\[c_{TL}(s) = \frac{g(s)}{g(s) + 1/256} \quad \forall \,\, s\in \{1,0\}^8.\]
\end{definition}

In the Continuation task, the table lookup prediction rule assigns to
every string $s \in \{1,0\}^7$ the empirical frequency with
which $s$ is followed by `1' in the training data. In the
Classification task, the table lookup prediction rule compares the empirical frequency of generation of string $s$ to the theoretical frequency of generation of string $s$ in a Bernoulli process.

Notice that the table lookup Continuation rule has $2^7$ free parameters
(corresponding to the $2^7$ unique strings of length seven), and the
table lookup Classification rule has $2^8$ free parameters
(corresponding to the $2^8$ unique strings of length eight). With over
21,000 observed strings, we have enough observations per unique string
to densely populate each cell of the lookup table. Thus, the table lookup prediction errors approximate the best possible prediction errors in these problems. See Appendix \ref{app:bounds} for more detail.

Table \ref{tab:comparison} reports the (tenfold cross-validated) prediction errors achieved by table lookup. These errors are then used as benchmarks against which to compare the prediction errors achieved using the behavioral models discussed above.
\begin{table}[h]%
	\caption{Comparison of prediction errors achieved using existing models with prediction errors achieved using table lookup. The behavioral models explain up to 15\% of the explainable variation in the data.}
	\label{tab:comparison}
	\begin{minipage}{\columnwidth}
\begin{center}
\bigskip
\begin{tabular}{ccccc}	
\hline
&\multicolumn{2}{c}{Continuation}& \multicolumn{2}{c}{Classification}\\
	& Error & Completeness &  Error& Completeness \\
	\hline
	Bernoulli & 0.25 & 0 & 0.25 & 0 \vspace{2mm}\\
Rabin (2002) &0.2494 & 0.10 &   0.2493 & 0.09\\
&(0.0007)& &(0.0008)\vspace{2mm}\\
Rabin \& Vayanos (2010) & 0.2492&0.13 &0.2488 & 0.15\\
&(0.0007)&&(0.0006)\vspace{2mm}\\
Table Lookup &0.2439 &1 &   0.2422 & 1 \\
&(0.0019) & &(0.0010) \vspace{2mm}\\
\hline
\end{tabular} 
\end{center}
		\centering
	\end{minipage}
\end{table}%

We find that table lookup achieves a prediction error of 0.2439 in the Continuation task and 0.2422 in the Classification task. The performance of table lookup is far worse than perfect prediction, showing that there is a large amount of irreducible noise in the problem of predicting human-generated coin flips. This emphasizes that naively comparing achieved prediction error to perfect prediction can, and in this case does, misrepresent the performance of the existing theories. 

A more appropriate notion of the achievable performance in this problem is the error achieved using table lookup. The errors of 0.2439 and 0.2422 above represent the predictive limits of the problems: conditioning only on initial flips, it is not possible to reduce prediction error  from the naive baseline by more than 0.0061 in Continuation and 0.0078 in Classification.  We propose as a simple measure of the {\em completeness} of the existing theories, then, the ratio of the reduction in prediction error achieved by the best behavioral model (relative to the naive baseline) to the reduction achieved by table lookup (again relative to the naive baseline). In the Continuation task, we find the completeness of the \citeasnoun{Rabin} and \citeasnoun{gambler} models to be up to 13\%, and in the Classification task, we find the completeness of these models to be up to 15\%.\footnote{For example, the completeness of the \citeasnoun{gambler} model in the Continuation task is computed as $(0.25 -  0.2492)/(0.25 - 0.2439) = 0.13$.}
 These results suggest that existing behavioral models produce between 13-15\% of the achievable improvement in prediction error.

As a robustness check, we repeat this exercise in Appendix \ref{app:supplementary} for different string lengths. For Continuation, this means using flips 1 through $k-1$ to predict the $k$-th flip, where $k$ varies from 2 to 7. For Classification, this means separating length-$k$ Bernoulli strings from the first $k$ flips generated by a human, where $k$ varies from 2 to 7. We find that prediction accuracy roughly increases in the length of the string (so that conditioning on a larger number of initial flips results in better prediction of the subsequent flip), but neither the errors nor the measures of completeness vary significantly for lengths near $k=8$.

\section{Features Versus Combination Rules}
The previous section found that human-constructed theories achieve only a fraction of the achievable performance gain over naive guessing. We ask now whether the limitations of existing behavioral models relative to table lookup arise because: (a) the behavioral models miss crucial predictive properties of the initial flips, or (b) they use the ``right" features, but do not combine them as effectively for prediction. 

To distinguish between these possibilities, we construct a feature space based on the existing models and related literature, and apply standard machine learning algorithms (Lasso regression and decision trees) to learn rules for combining these known features. We find that these algorithms predict significantly better than the behavioral models, and in fact closely approximate the performance of table lookup. Moreover, a substantial amount of the improvement over the behavioral models persists even when these scalable algorithms are restricted to use of only a small number of features (comparable to the number of free parameters in the behavioral models). This suggests that the reason in (b) accounts for at least a part of the gap between the behavioral models and table lookup: alternative models based on similar features can substantially improve performance. 

We then ask whether the use of additional properties of the initial flips, not yet captured in existing models, can further improve predictive performance. Towards this goal, we define a rich set of ``atheoretical" features. Each binary feature corresponds to a possible substring pattern, and takes value `1' when the pattern appears in the initial flips. We use machine learning algorithms to discover the most predictive features from this rich set of patterns, and then predict based on combinations of these features. Use of the decision tree algorithm with this feature space represents a compression of the table lookup predictor\textemdash instead of assigning a prediction to each unique binary string, it partitions the space of strings, and learns a constant prediction for each partition element.

We find that prediction rules based on behavioral features are substantially more predictive than prediction rules based on an equal number of (best) atheoretical patterns. Moreover, when we combine the set of features, but continue to impose a restriction on the number of features, we find that only behavioral features are selected by the algorithms. This suggests that the behavioral features are more predictive than the best (small) subset of atheoretical patterns.\footnote{This comparison is not precise: the atheoretical features are binary-valued, while the behavioral features take values from a larger set, so the information content in the latter features is intrinsically larger.}

These results collectively suggest that gap between table lookup and
the behavioral models in our domain
is better explained by (b) than (a)\textemdash
the research community approximately knows the ``right" features for
the problem, but may not combine them as effectively as the machine
learning algorithms.

\subsection{Prediction Rules Based on Behavioral Features} \label{sec:pred-behav}
We begin by constructing a feature space based on the relevant literature, including the features:
\begin{itemize}
\item the proportion of alternation in the string (averaged across all flips)
\item the total number of runs of length $k$ in the string (for Classification, we allow $k$ to vary from 2 to 8, and for Continuation, we allow $k$ to vary from 2 to 7)
\item the number of Heads in the string
\item the length of the longest run at the beginning of the string
\item the length of the longest run at the end of the string
\end{itemize}
and all of their pairwise interactions. This makes for 55 features in
the Continuation task, and 66 features in the Classification task.
Every binary string is recoded as a feature vector for each of the
prediction tasks, so that prediction rules are maps from feature
vectors to probabilities.\footnote{In the Continuation task, a
prediction rule is a map from the set of feature vectors describing
length-7 strings to a probability that the final flip is Heads. In
the Classification task, a prediction rule is a map from the set of
feature vectors describing length-8 strings to a probability that the
string was generated by a human.}

We use two standard machine learning algorithms\textemdash Lasso
regression and decision trees (see e.g. \citet{Hastie})\textemdash to
select a prediction function based on the above features. From Table
\ref{tab:ML}, we see that the cross-validated prediction errors
obtained using these approaches closely approximate the table lookup
prediction errors. In both problems, the best machine learning
algorithm achieves approximately 80-90\% of the achievable reduction in
prediction error. This means that for our domain, there is relatively
little loss in using the
best scalable machine learning algorithm (trained on behavioral
features) as a substitute benchmark for table lookup.

\begin{table}[H]%
	\caption{The performance of scalable algorithms approximates
	table lookup.} \label{tab:ML} \bigskip
	\begin{minipage}{\columnwidth}
\begin{center} \begin{tabular}{ccc|ccc} \hline &
\multicolumn{2}{c}{Continuation} & \multicolumn{2}{c}{Classification}
\\[1mm] \hline
 &  Error &  Completeness &   Error &  Completeness \\ \hline
    Naive &0.25  & 0 & 0.25 & 0  \\[2mm] Lasso & 0.2475 & 0.41 &
    0.2444 & 0.72 \\
	&\footnotesize{($0.0007$)}& & \footnotesize{($0.0003$)} \\
    Decision Tree & 0.2443 & 0.93  & 0.2437 & 0.81 \\
    &\footnotesize{($<$0.0000)}& &  \footnotesize{($<$0.0000)} \\
	Table Lookup &0.2439    & 1 & 0.2422 & 1 \\
&\footnotesize{(0.0019)} & &\footnotesize{(0.0010)}\\
    \hline
\end{tabular} \end{center}

	\centering
	\end{minipage}
\end{table}%

\noindent These results suggest that new combinations of known features can yield large improvements in prediction. We turn next to considering whether combinations of \emph{small} numbers of known features can also yield large improvements in prediction. 
 
\subsection{Restriction to a Small Number of Features} \label{sec:small}
We again train prediction rules using the set of features described above, under a new constraint on the number of parameters. Below, we show the prediction errors obtained by decision trees that are restricted to $k$ splits, where we consider $k=2$ (thus comparing to the behavioral models), $k=3$ and $k=5$. (The $k=2$ split decision tree can be found in Figures \ref{fig:BehavCont2} in the appendix.)

\begin{table}[H]%
	\caption{Machine learning algorithms built on behavioral features predict well even when restricted to use of a small number of features.}
	\label{tab:MLshort}
	\bigskip
	\begin{minipage}{\columnwidth}
\begin{center}
\begin{tabular}{ccc|ccc}
\hline
& \multicolumn{2}{c}{Continuation} & \multicolumn{2}{c}{Classification} \\[1mm]
\hline
 &  Error &  Completeness &   Error &  Completeness \\
 \hline
    Naive &0.25   & 0 & 0.25 & 0  \\[2mm]

            2 parameters & 0.2477 & 0.38  & 0.2459 & 0.53 \\
    &\footnotesize{($<$0.0000)}& &  \footnotesize{($<$0.0000)} \\
                3 parameters & 0.2470 & 0.49  & 0.2457 & 0.55 \\
    &\footnotesize{($<$0.0000)}& &  \footnotesize{($<$0.0000)} \\
                5 parameters & 0.2461 & 0.64 & 0.2451 & 0.63 \\
    &\footnotesize{($<$0.0000)}& &  \footnotesize{($<$0.0000)} \\
            Table Lookup &0.2439   & 1 & 0.2422 & 1 \\
 &\footnotesize{(0.0019)} & &\footnotesize{(0.0010)}\\
    \hline
\end{tabular}
\end{center}

	\centering
	\end{minipage}
\end{table}

With two parameters, the best decision tree achieves 38-53\% of the possible reduction in prediction error, and with five parameters, the best decision tree achieves up to 64\% of the possible improvement. These results suggest that scalable algorithms can attain a substantial improvement on prediction error even when restricted to use of a very small number of features. (Recall for comparison that the best behavioral models achieved up to 15\% of the table lookup improvement in these problems.)

Previously, we considered use of the best \emph{unrestricted} machine learning algorithm as a substitute benchmark for table lookup. If we instead construct a benchmark using the performance of the best two-parameter decision tree\textemdash thus, comparing the existing models against a relatively interpretable model\textemdash we find that the completeness of the behavioral models is 35\% in the Continuation task and 30\% in the Classification task.\footnote{Continuation: $(0.25-0.2492)/(0.25-0.2477)=0.35$; Classification: $(0.25-0.2488)/(0.25-0.2459)=0.29$.} So, conditioning on a small number of interpretable features only, there is still room for improvement.

\subsection{Comparison with Algorithmic Features} \label{sec:algorithmic}

We turn next to the question of whether it is possible to improve upon
the performance of the algorithms above by discovering new features
from a rich set of algorithmic patterns. 
To study this, we build a second feature
space using features corresponding to the presence
of specific patterns in the sequences of coin flips.
We define a {\em pattern} $p$ to be 
a length-$k$ binary sequence over the alphabet
$\{1,\ast\}$, and we say that the length-$k$ string $s \in \{1,0\}^k$
{\em contains} the length-$k$ pattern $p \in \{1,\ast\}^k$ if
$s_i = p_i$ at every index $i$ where $p_i=1$.
Thus, for example, the string 1011011 contains the patterns 
1$\ast$1$\ast$$\ast$$\ast$$\ast$ and $\ast$$\ast$$\ast$$\ast$$\ast$11,
but not the pattern 11$\ast$$\ast$$\ast$$\ast$$\ast$.

We construct this second feature space by
enumerating each substring pattern $p$ (where $p \in
\{1,\ast\}^7$ for the Continuation task and $p \in \{1,\ast\}^8$ for the
Classification task), and defining indicator variables for the appearance
of that pattern in each possible input string $s$
(where $s \in \{1,0\}^7$ for Continuation $s \in \{1,0\}^8$ for Classification).
We allow machine learning algorithms to select the $k=2$, 3,
and 5 most useful such features for prediction from among this set 
of pattern indicators, 
and we show the resulting prediction errors 
in Table \ref{tab:alg}.
Again, the decision
trees for $k=2$ can be found in the appendix (see Figure
\ref{fig:AlgCont2}).

\begin{table}[H]%
	\caption{Prediction rules using a small number of algorithmic features perform worse than prediction rules using the same number of behavioral features.}
	\label{tab:alg}
	\bigskip
	\begin{minipage}{\columnwidth}
\begin{center}
\begin{tabular}{ccc|ccc}
\hline
& \multicolumn{2}{c}{Continuation} & \multicolumn{2}{c}{Classification} \\[1mm]
\hline
 &  Error &  Completeness &   Error &  Completeness \\
 \hline
    Naive &0.25   & 0 & 0.25 & 0  \\[2mm]

            2 parameters & 0.2488 & 0.20 & 0.2492 & 0.10 \\
    & \footnotesize{(0.0001)}& &  \footnotesize{(0.0001)} \\
              3 parameters & 0.2483 & 0.28  & 0.2491 & 0.11 \\
    & \footnotesize{(0.0001)}& &  \footnotesize{(0.0001)} \\
                5 parameters & 0.2479 & 0.34 & 0.2486 & 0.18\\
    & \footnotesize{(0.0001)}& &  \footnotesize{(0.0001)} \\
            Table Lookup &0.2439   & 1 & 0.2422 & 1 \\
&\footnotesize{(0.0019)} & &\footnotesize{(0.0010)}\\
    \hline
\end{tabular}
\end{center}

	\centering
	\end{minipage}
\end{table}%

Table \ref{tab:alg} shows that these prediction errors are higher than
the corresponding errors in Table \ref{tab:MLshort} for every
prediction task and restriction on number of parameters that we
consider. These results provide further evidence that the gap between
table lookup and the behavioral models does not necessarily
imply that researchers are missing key predictive features\textemdash
searching over a space of new syntactic patterns for strings does
not produce improvements over the combinations of known features that we considered in Section \ref{sec:pred-behav}.

\section{Robustness of the Benchmark} \label{sec:robustness}
In Section \ref{sec:main}, we introduced table lookup as a means of approximating the best possible prediction error. Relative to the table lookup performance, we found that existing models achieve
approximately 13-15\% of the achievable improvement in prediction error. Because table lookup is extremely flexible, however, it is possible that it learns a highly precise, but highly specific, model of human generation of coin flips of length-8. For example, it turns out that 57\% of the strings in our dataset begin with ``Heads."  Table lookup learns this (domain-specific) asymmetry
and uses it for prediction, but the probability of `1' in the first
flip is fixed to be 0.5 in both the \citet{Rabin} and \citet{gambler}
models. This raises the question of to what extent the improvement of table lookup over existing models can be attributed to idiosyncratic features related to generation of length-8 coin flips.

To test the specificity of the table lookup predictor to our problem domain, we ask how well this predictor can transfer predict strings generated in a different but neighboring domain. We focus on \emph{small} changes in the prediction domain, across which we would expect the behavioral models to be stable. If indeed table lookup adapts sensitively to fine details of the original context, then it may fail to predict even strings generated under these minor changes in framing.  This kind of ``conceptual overfitting" represents the possibility for a richly-parametrized model to generalize poorly not because of insufficient data \emph{within-domain}, but because of an insufficient sampling of contexts across which the parameters vary. 

The new problem domains we consider are the following: first, we change the \emph{alphabet} from which
the realizations are drawn; second, we change the \emph{length} of the flips to be predicted; finally, we consider prediction of the strings generated in the work of 
\citet{distributional}, which were produced under different conditions
and by a different subject pool from our main dataset. For each of these, we train
prediction rules on the original data of length-8 coin flips, and then
ask how well these rules predict strings that are generated under the
new framing.  We find that the transfer performance of table lookup is
comparable to its ``within-domain" performance in the previous
section, in which the test and training data were generated from the
same context. Moreover, the measure of completeness is also stable
across local problem domains: existing models produce no more than 20\%
of the improvement in prediction error obtained by using table lookup
for transfer prediction.

\subsection{New Datasets}
We briefly describe below the data collected in each of the new domains.

\emph{New alphabet.} The first transfer domain re-labels the outcome space from  $\{H,T\}$  to $\{r,2\}$. We asked 124 subjects on Mechanical Turk to generate 50 binary strings of length eight as if these strings were the realizations of 50 experiments in which a fair coin labelled `$r$' on one side and `2' on another was flipped 8 times. This yielded a total of 6,200 strings. As in the main experiment, subjects were given only 30 seconds to complete each string, and payment was conditioned on whether their strings looked plausibly random.  These properties of the experimental design are consistent across all of the experiments, so we will not repeat them below. 

\emph{New length.} The second domain changes the string length from eight to 15. We asked 120 subjects on Mechanical Turk to generate 25 binary strings of length fifteen as if these strings were the realizations of 25 experiments in which a fair coin was flipped 15 times. This yielded a total of 3,000 strings. We use these length-15 strings to construct seven datasets of length-8 strings, each including only flips $k$ through $k+7$, where $k \in \{2,\dots,8\}$.\footnote{We leave out $k=1$, which would correspond to the original setting.}

\emph{New subject pool.}  The final domain considers prediction of data from \citet{distributional}, in which thirty Tufts undergraduates were asked to each produce 100 binary sequences, ``as if 100 people had each tossed a coin 10 times, and the results had been recorded in a table of 100 rows and 10 columns, with each row corresponding to an individual." We use a public version of their data that includes the responses of 28 subjects and a total of 2,800 strings. As above, we construct truncated datasets of length-8 strings, including flips $k$ through $k+7$, this time taking $k=1,2,3$. 

\begin{figure}[h]
\begin{center}
\includegraphics[scale=.32]{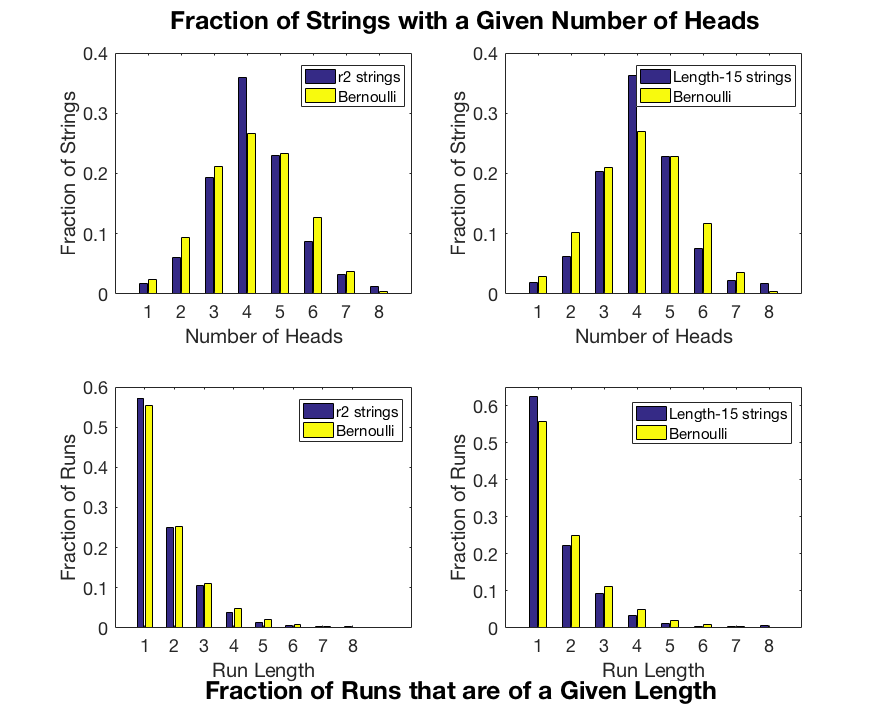}
\end{center}
\caption{\footnotesize{(a) Top row:  the fraction of strings that include at least one run of length $m$, where $m$ is the label on the $x$-axis. \textit{Left}\textemdash comparison of $\{r,2\}^8$ data (purple) with simulated Bernoulli strings (yellow); \textit{Right}\textemdash comparison of the $\{H,T\}^{15}$ data (purple) with simulated Bernoulli strings (yellow). (b) Bottom row: the fraction of generated strings that include $m$ Heads, where $m$ is the label on the $x$-axis. \textit{Left}\textemdash comparison of $\{r,2\}^8$ data (purple) with simulated Bernoulli strings (yellow); \textit{Right}\textemdash comparison of the $\{H,T\}^{15}$ data (purple) with simulated Bernoulli strings (yellow).}}
\label{fig:sumTransfer}
\end{figure}

Below, we briefly compare summary statistics of these new datasets with the original coin flip data. The basic distributional facts are similar: in the original coin flip data, Heads was produced in 52.61\% of flips; under the new framings, the symbol `r' is produced in 50.91\% of flips in the $\{r,2\}^8$ data, and Heads is produced in between 50.53\%-51.73\% of flips in the different cuts of the $\{H,T\}^{15}$ data.  The features of misperception of randomness discussed previously in Section \ref{sec:data} appear also in the new data. As we show in Figure \ref{fig:sumTransfer}, subjects under-generate long runs and over-generate balanced strings in all of these settings.\footnote{The fractions of strings shown for the $\{H,T\}^{15}$ data are averaged across the seven datasets of flips $k$ through $k+7$. The comparison Bernoulli distributions are found by simulating a dataset for each of $k=2, \dots,8$, where the probability of Heads is set to the mean flip in the dataset of strings $k$ through $k+7$.}

\subsection{Transfer Prediction} \label{sec:transfer}

In what follows, we build a table lookup predictor from the original coin flip data described in Section \ref{sec:data}, and use this to predict strings generated in the related domains described above. For prediction of strings from $\{r,2\}^8$, the transfer Continuation task is to predict the final outcome from $\{r,2\}$ given the first seven, and the transfer Classification task is to separate the human-generated strings from $\{r,2\}^8$ from an equal number of Bernoulli strings. For prediction of substrings from the $\{H,T\}^{15}$ data, we treat each set of truncated strings as a distinct dataset, and define separate transfer Continuation and Classification problems for each cut. The prediction errors we report in these problems are averaged across the seven truncated datasets.

Throughout, we use the arbitrary convention that `H' (in the original coin flip
data) is identified with `r' in the first transfer problem, and with
`H' in the second.\footnote{Thus, the strings `HTHTHTHT' and
`r2r2r2r2' are identically coded as `10101010'.} Prediction errors
obtained assuming the reverse mapping are presented in Tables
\ref{tab:Flippedcont}, \ref{tab:Flippedclass}, and \ref{tab:FlippedNB} 
of the appendix, and
do not differ substantially from those shown below. All errors
presented in Tables \ref{tab:Transfercont} and \ref{tab:Transferclass}
are mean-squared errors with bootstrapped standard errors.

Since these are transfer prediction problems, it is possible for the prediction rules we consider to perform worse than guessing at random.\footnote{For example, it may be that while strings exhibit negative autocorrelation in our primary setting, they exhibit positive autocorrelation in one of the new prediction domains.} This turns out not to be the case: across each of the new prediction domains, the prediction rules we consider are weakly predictive. Additionally, the prediction errors are comparable in absolute terms across domains: for example, table lookup achieves prediction errors ranging from 0.2325 to 0.2434 in these transfer problems. This interval is similar to the previous errors of 0.2422 and 0.2439 in the original task.\footnote{The lower prediction errors in the transfer tasks reflect settings in which there is less randomness in the generated strings (to be predicted).} Finally, the measures of completeness that we find in these transfer problems do not improve significantly upon the within-domain estimates, as we would expect them to if table lookup were indeed much less robust than the existing models. In these transfer exercises, the completeness measures do not exceed 20\%. These results suggest that the table lookup benchmark and the estimated measures of completeness do generalize across local changes in the problem domain.

\begin{table}[H]%
	\caption{Prediction errors in the \emph{transfer Continuation problem}\textemdash how well do prediction rules, trained on the original $\{H,T\}^8$ data, predict strings that are generated in a different but neighboring domain?}
	\label{tab:Transfercont}
	\bigskip
	\footnotesize
	\begin{minipage}{\columnwidth}
\begin{center}
\begin{tabular}{ccccccc}	
\hline
&\multicolumn{2}{c}{$\{r,2\}^8$}& \multicolumn{2}{c}{$\{H,T\}^{15}$} & \multicolumn{2}{c}{N-B (2009)} \\
	& Error & Completeness &  Error& Completeness  &  Error& Completeness \\
	\hline
	Naive & 0.25 & 0 & 0.25 & 0 &0.25 & 0 \\[2mm]
Rabin (2002) &0.2497 & 0.05 &    0.2489 & 0.08 &  0.2474 & 0.15\\
&(0.0005)&& (0.0005) && (0.0010) \\[2mm]
R-V (2010) & 0.2496& 0.06 &0.2472  & 0.20 & 0.2491 & 0.05 \\
&(0.0004)&&(0.0005) && (0.0003)   \\[2mm]
Table Lookup &0.2434 & 1  &0.2361 & 1 & 0.2325 & 1\\
&(0.0011) &&(0.0018) & & (0.0023) \\
\hline
\end{tabular} 
\end{center}
		\centering
	\end{minipage}
\end{table}%

\begin{table}[H]%
	\caption{Prediction errors in the \emph{transfer Classification problem}\textemdash how well do prediction rules, trained on the original $\{H,T\}^8$ data, predict strings that are generated in a different but neighboring domain?}
	\label{tab:Transferclass}
	\footnotesize
	\bigskip
	\begin{minipage}{\columnwidth}
\begin{center}
\begin{tabular}{ccccccc}	
\hline
&\multicolumn{2}{c}{$\{r,2\}^8$}& \multicolumn{2}{c}{$\{H,T\}^{15}$} & \multicolumn{2}{c}{N-B (2009)}\\
	& Error & Completeness &  Error& Completeness  &  Error& Completeness \\
	\hline
	Naive & 0.25 & 0 &0.25 & 0 &0.25 & 0 \\[2mm]
Rabin (2002) &0.2498 & 0.02 &    0.2490 & 0.10  & 0.25 & 0\\
&(0.0005)& &(0.0015) && (0.0006)\\[2mm]
R-V (2010) & 0.2495& 0.06 &0.2479 & 0.19 &  0.2497 & 0.02 \\
&(0.0003)&&(0.0006) && (0.0004) \\[2mm]
Table Lookup &0.2415 & 1 &  0.2392  & 1 & 0.2377 & 1\\
&(0.0003) &&(0.0013) && (0.0015)   \\
\hline
\end{tabular} 
\end{center}
		\bigskip\centering
	\end{minipage}
\end{table}%

\section{Application of Approach in Field Domains}

In what follows, we consider application of the proposed approach to testing
theory completeness in two field domains.  Our first field study uses
data from \citet{Shue}, and consists of the sequential judgments made
by baseball umpires regarding whether to call a pitch a {\em strike.} 
Our second
field study uses data from \citet{RPS} and consists of the repeated
decisions by Facebook players in online games of rock-papers-scissors. We
represent the umpire data as binary sequences, where the outcome is
whether the pitch is called as a {\em strike} or a {\em ball}. We represent the
Rock-Paper-Scissors data as ternary sequences of sequential throws,
where the outcome corresponds to whether the player chose rock, paper, or
scissors. In these settings, we can study questions that closely
resemble those considered in this paper so far: Given an initial set of
calls by an umpire (or throws in Rock-Paper-Scissors), can we predict
subsequent calls (throws)?  And, as a classification task,
given a set of strings, half of which
correspond to sequences of umpire calls (or Rock-Paper-Scissors throws), 
and half of which
correspond to independent realizations of a random variable, can we
determine the source of generation?

We find that table lookup provides a non-trivial
improvement over naive guessing in both of these domains, and
can again be used to construct a benchmark for
attainable predictive accuracy in these problems. 
Interestingly, in both field domains, the model of \citet{gambler} achieves
levels of predictive gain relative to
this table lookup benchmark that are 
qualitatively similar to what we observed in our 
experimental data in Section \ref{sec:benchmark}.
This suggests that the completeness of this theory is 
relatively stable despite wide differences in the nature of the domains.

\subsection{Datasets}
We now describe each of these datasets in greater detail.

\emph{Baseball Umpires.} An important role of baseball umpires is to determine whether pitches should be called as {\em balls} or {\em strikes} when the batter does not swing. A designated strike zone takes the shape of a vertical right pentagonal prism located above home plate, and the umpire should call ``strike" whenever the ball is within the strike zone as it passes the location of home plate, and ``ball" otherwise. While the definition of a strike is objective, the judgment of whether or not the pitch constitutes a strike is not.

\citet{Shue} collected a large dataset of umpire calls, including approximately 1.5 million pitches over 12,564 games by 127 different umpires. We extract from this dataset non-overlapping sequences of consecutive calls of length 6 that occurred within the same game. Our data includes a total of 15,127 strings, where the frequency of strikes (across all pitches) is approximately 27\%, and the mean frequency of strikes in the final call is 33\%. \citet{Shue} find that these calls exhibit negative auto-correlation: in aggregate, umpires tend to avoid  runs of the same call (i.e. calling many strikes in a row or many balls in a row). Our goal is to exploit further context-dependence towards prediction. 

\emph{Rock-Paper Scissors.} A Facebook app called Roshambull allowed Facebook users to play games of Rock-Paper-Scissors against one another throughout the year 2007. Each game consisted of two players and lasted until either player had won two matches, where the two winning matches need not be consecutive. 

\citet{RPS} collected a dataset of 2,636,417 matches, all of which occurred between May 23, 2007 and August 14, 2007.\footnote{In these games, players were shown the history of their opponents' previous plays. The way in which this information influenced play is a focus of \citet{RPS}, but we do not address it.} We extract from this dataset the initial consecutive six choices in all games that lasted at least six matches, and consider each of these a string.\footnote{The average game lasted 4.29 matches.}  Our data includes a total of 29,864 strings, where the overall frequency of Rock is 37.42\%, Paper is 33.58\%, and Scissors is 29.00\%. Unlike the other domains studied in this paper, the Rock-Paper-Scissors strings exhibit positive autocorrelation: the probability that a throw is followed by a different throw is 0.64, which is slightly less than the expected level of alternation given independent throws. When we apply \citeasnoun{gambler} for prediction, we therefore relax the constraint that the free parameters $\alpha, \delta$ have positive values, so that the model serves as a general model of autocorrelated strings. 

Within these new domains, where strings are generated by people making sequential decisions in a real environment, we return to the question of the previous sections: how well can we predict individual entries in
these human-generated strings, and how well can we distinguish them
from Bernoulli strings with corresponding parameters?

\subsection{Establishing a Benchmark} \label{sec:field}
Following the approach outlined in Section \ref{sec:main}, we use table lookup to construct a benchmark for the achievable level of prediction in these new field domains. 

In the umpire setting, the objects of prediction are binary strings of length 6. The Continuation task in this domain is to predict the final flip given the first five, and the Classification task is to separate strings of umpire calls from synthetic Bernoulli strings, with the probability of `1' set to equal the average flip in the umpire data (0.27). As before, the naive prediction error in the Classification problem is found by predicting 0.50 unconditionally. Because of the asymmetry in `1's and `0's, the naive prediction rule in the Continuation problem is different: we learn the average final flip in the training data, and predict this average unconditionally in the test data. These naive rules yield prediction errors of 0.2213 in the Continuation task and 0.25 in the Classification task.\footnote{Notice that the naive prediction error in Continuation is less than 0.25 because of the asymmetry of `1's and `0's.}

Using \citet{gambler} to predict in this setting, we find
cross-validated prediction errors of 0.2204 in the Continuation task
and 0.2489 in the Classification task. Thus, as in our main setting,
the behavioral models are predictive.\footnote{In the Classification
task, the improvement in prediction error is statistically
significant, although it is not for Continuation.} We ask next how
complete they are, and answer this again by using table lookup to
establish a benchmark for our notion of
the best achievable predictive accuracy in
this domain. We find that the table lookup prediction errors are
0.2171 in the Continuation task and 0.2433 in the Classification task.
Comparing the improvement upon the naive baseline achieved by the
\citet{gambler} prediction rule, and by the table lookup rule, we find
that \citet{gambler} achieves 17-21\% of the attainable improvement in
this problem. Robustness checks are reported in Appendix \ref{app:field}, in
which we vary the length of the strings, predicting instead strings of
length 5 and length 4.

\begin{table}[t]%
	\caption{Predicting umpire calls: \citet{gambler} explains 17-21\% of the explainable variation in the data.}
	\label{tab:umpire}
	\bigskip
	\begin{minipage}{\columnwidth}
\begin{center}
\begin{tabular}{ccccc}	
\hline
&\multicolumn{2}{c}{Continuation}& \multicolumn{2}{c}{Classification}\\
	& Error & Completeness &  Error& Completeness \\
	\hline
	Naive &  0.2213\footnote{The standard error of the above estimate is 0.0030.} & 0 &0.25 &0 \vspace{2mm}\\
Rabin and Vayanos (2010) &  0.2204 & 0.21 &0.2489 & 0.17\\
&(0.0044)&&(0.0005)\vspace{2mm}\\
Table Lookup &   0.2171 & 1 &   0.2433 & 1\\
&(0.0039) &&(0.0024) \vspace{2mm}\\
\hline
\end{tabular} 
\end{center}
		\centering
	\end{minipage}
\end{table}%

Turning now to the Rock-Paper-Scissors data, the objects of prediction become \emph{ternary} strings of length 6. The Continuation task in this domain is to predict the final throw given the first five, and the Classification task is to separate strings of Rock-Paper-Scissors throws from synthetic strings of length 6, where each element is randomly drawn from $\{r,p,s\}$. In the former problem, a prediction rule is a map from strings in $\{r,p,s\}^5$ to probability vectors in $ [0,1]^3$, where each coordinate corresponds (in order) to the probability of realization of $r$, $p$, or $s$. The realized throw is represented as a binary vector of length 3,  which takes value `1' in the coordinate corresponding to the throw in observation $i$ (so that for example, throw of Rock is represented by $(1,0,0)$). We consider the loss function 
\[\frac1n \sum_{i=1}^n \| \bold{y}_i- \bold{q}_i\|_2,\]
where $\bold{q}_i$ is the predicted probability vector for observation $i$ and $\bold{y}_i$ is the outcome. Naively guessing a probability of $1/3$ for each outcome yields a prediction error of $0.8165$ in the Continuation task, and guessing $1/2$ unconditionally yields a prediction error of $0.25$ in the Classification task. 

Using the relaxed \citeasnoun{gambler} prediction model (which allows $\alpha, \delta <0$), we obtain a cross-validated prediction error of 0.8160 in the Continuation task and 0.2491 in the Classification task. Thus, these models are again predictive. Turning to the question of how complete they are, we find that the table lookup errors are 0.8129 in the Continuation task and 0.2417 in the Classification task. Comparing the improvement upon the naive baseline achieved by the \citet{gambler} prediction rule, and by table lookup, we find that \citet{gambler} achieves 6-11\% of the attainable improvement in this problem. Again, see Appendix \ref{app:field} for robustness checks in which we predict instead strings of length 5 and length 4.

\begin{table}[H]%
	\caption{Predicting Rock-Paper-Scissors throws: \citet{gambler} explains 6-11\% of the explainable variation in the data.}
	\label{tab:RPS}
	\bigskip
	\begin{minipage}{\columnwidth}
\begin{center}
\begin{tabular}{ccccc}	
\hline
&\multicolumn{2}{c}{Continuation}& \multicolumn{2}{c}{Classification}\\
	& Error & Completeness &  Error& Completeness \\
	\hline
	Naive & 0.8165 & 0 & 0.25 & 0 \vspace{2mm}\\
Rabin and Vayanos (2010) &      0.8160   & 0.06 & 0.2491 & 0.11\\
&(0.0005)&&(0.0003)\vspace{2mm}\\
Table Lookup & 0.8129 & 1 &  0.2417 & 1\\
& (0.0019) &&(0.0013) \vspace{2mm}\\
\hline
\end{tabular} 
\end{center}
		\centering
	\end{minipage}
\end{table}%

Notice from Tables  \ref{tab:umpire} and \ref{tab:RPS} that the measures of completeness in these field settings (which vary from 6 to 21\%) do not differ substantially from the measure of completeness elicited using the experimental data in Section \ref{sec:benchmark} (approximately 15\%). This suggests that the extent to which the \citeasnoun{gambler} model captures predictable structure is approximately stable across the different domains we have considered. Notice that this stability holds despite substantial differences in the nature of these domains: the experimental setting can be understood as one of pure generation, whereas the baseball umpire setting involves inference about an underlying state, and the Rock-Paper-Scissors setting involves beliefs about the random sequences generated by other players.

Although the completeness measure is stable within a range, the
variations across the settings are revealing. Compare, for example,
the table lookup prediction errors attained in the Classification
problem in each of the domains.\footnote{Comparisons of absolute
levels in the Continuation task are harder to interpret, since the
average flip varies across each of the domains.} This error is 0.2433
in the baseball umpire setting, 0.2422 in the experimental setting,
and 0.2419 in the Rock-Paper-Scissors setting. Recalling that the
table lookup prediction error represents the lowest achievable
prediction error, this comparison suggests that there is less
predictable structure in the sequences of baseball umpire calls,
relative to the other two domains.\footnote{An interesting
feature of the baseball domain --- and a contrast with our other domains ---
is that there is a ``true'' answer for each pitch (ball or strike)
that the umpire is trying to judge as accurately as possible.
As a result, the sequence of umpire calls depends both on this sequence
of true answers, and on the sequence of umpire judgments.
Since the data of \citet{Shue}
contained the true answer for each pitch,
we investigated the extent to which table lookup could be used to 
predict this true answer from the sequence of prior true answers
(independently of the umpire's call), and we found that for this
Continuation task on true answers there is very little
predictable structure in the sequence.  Thus, the prediction performance
on the sequence of umpire calls obtained by table lookup in this 
domain (at least for the Continuation task) appears to be coming 
primarily from the stucture in the sequence of calls themselves,
rather than from the sequence of true answers.  This is consistent
with analysis of \citet{Shue}, who argued that the auto-correlation
in the sequence of calls arises principally from the umpire's decisions.
}

Additionally, the contrast between Tables \ref{tab:umpire} and \ref{tab:RPS} shows that the model based on \cite{gambler} is a more complete predictor of umpire calls than of Rock-Paper-Scissors throws. Although this difference is intuitive \textemdash \, sequences of umpire calls are approximately a direct test of the \citeasnoun{gambler} model, while Rock-Paper-Scissors involves (un-modeled) strategic considerations \textemdash \, illustration of this difference requires a notion such as our proposed measure of completeness. In particular, notice that the \citeasnoun{gambler} model is not only predictive in both domains, but reduces the naive prediction error by similar margins. A straightforward comparison of prediction errors, therefore, does not reveal that the improvement is more substantial in one domain than the other.  Having table lookup as a baseline in both settings
makes it possible to demonstrate this and to quantify the difference.

\section{Discussion}
When evaluating the predictive performance of a theory, it is important to know not just whether the theory is predictive, but also how complete its predictive performance is. To assess theory completeness, we need a notion of what constitutes the best \emph{achievable} predictive performance for a given problem. This is difficult to assess in general, but we introduce a social science domain\textemdash human perception and generation of randomness\textemdash in which it is possible to search the space of predictive models to optimality. This permits construction of a benchmark for the best achievable level of prediction, which we use to evaluate the predictive performances of leading economic theories in the domain: We find that these theories explain roughly 13-15\% of the explainable variation in experimental data, and show moreover that table lookup can be used to construct benchmarks for prediction of field data that correspond to natural instantiations of human generation of randomness.

The table-lookup approach also provides a conceptual framework for thinking about how the best possible predictive accuracy changes with the set of measured features. Consider a comparison between the lookup table we currently have, with columns corresponding to the features $\{x_i\}$, and a larger lookup table in which we add a new set of currently unmeasured features $\{z_j\}$, with a new column for each of these new features. With sufficient data, the larger table provides (weakly) better predictive accuracy than the original smaller table, and the limitations in using the smaller table for prediction can be thought of as the consequence of implicitly averaging over all the possible values for the unmeasured features. In the context of the generation and perception of randomness, it becomes interesting\textemdash both as a goal in itself and as a perspective on the limitations of our current feature set\textemdash to consider what might constitute a set of unmeasured features of the human participant's behavior that may improve predictive accuracy if we chose to add them as columns to the table.

Two additional future directions are of interest: first, are there
other social science problems in which brute force techniques such as
table lookup might apply? Second, when brute force techniques such as
table lookup are not be feasible, can we still use approaches from
machine learning to construct a benchmark for the attainable level of
prediction, and how do such approaches perform in practice? 
Our analyses here, together with 
results in the contemporaneous work of \citet{Naecker}, suggest that
``approximate" benchmarks based on scalable machine learing algorithms 
may be effective practical solutions.

\clearpage
\appendix

\clearpage
\pagebreak

\section{Supplementary Materials to Section \ref{sec:data} (Experimental Instructions)} \label{instructions}
Subjects on Mechanical Turk were presented with the following introduction screen:
\begin{figure}[H]
\begin{center}
\label{heads1}
\includegraphics[scale=0.48]{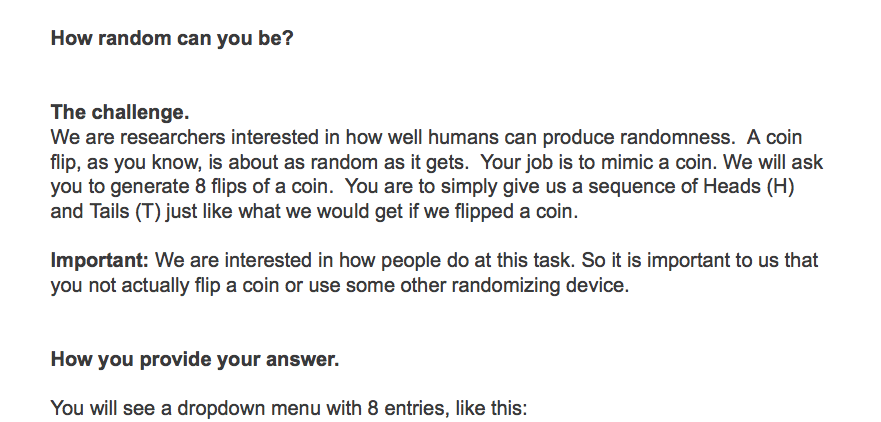}
\includegraphics[scale=0.48]{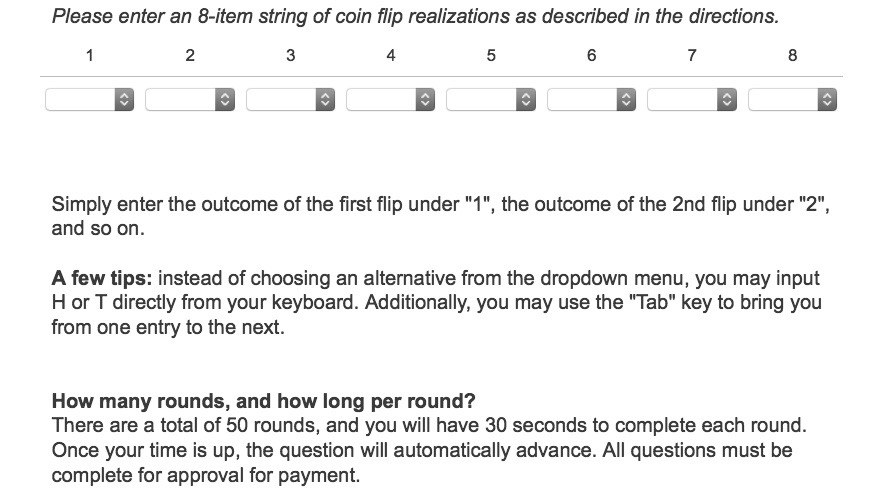}
\includegraphics[scale=0.48]{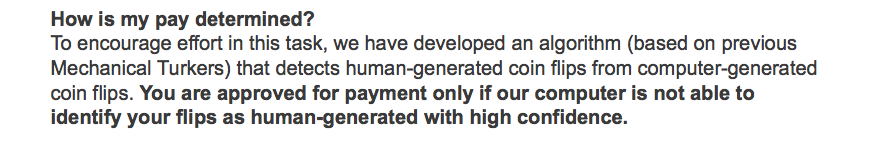}
\end{center}
\end{figure}

\noindent Following a trial round and statement of consent, subjects were presented with 50 identical screens that looked like the following:
\begin{figure}[H]
\begin{center}
\label{heads2}
\includegraphics[scale=0.5]{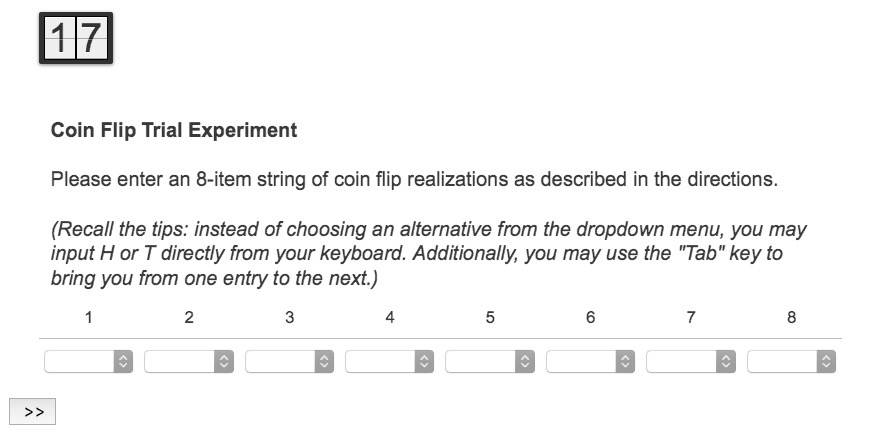}
\end{center}
\end{figure}
\noindent Subjects were given 30 seconds to complete each string, and a timer displayed their remaining time. 

Other versions of the experiment used similar instructions with small variations. For example, the experiment eliciting length-8 strings from $\{r,2\}^8$ used the instructions:
\begin{figure}[H]
\begin{center}
\label{heads3}
\includegraphics[scale=0.35]{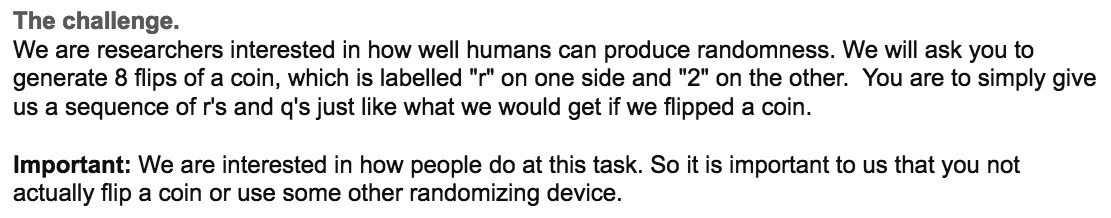}
\end{center}
\end{figure}

\clearpage
\pagebreak

\section{Supplementary Materials to Section \ref{sec:behavioral-models} (Behavioral Prediction Rules)} \label{appendix:predrule}
	
The prediction rules based on \citet{Rabin} include free parameters $p \in [0,1]$ and $N \in \mathbb{N}$. For Continuation, the probability that string $s \in \{1,0\}^7$ is followed by `1' is predicted to be
\[f_R(s)= 0.5 p + \sum_{k=0}^6 p(1-p)^k \left(0.5N - \textstyle \sum_{j=7-k}^7 s_k\right)/N.\]
where $s_k$ is the $k$-th flip in string $s$. For Classification, first define
\[q(s|r) = 0.5 \cdot r_k + \frac{1-r_k}{N} \left( 0.5N - \textstyle \sum_{j=1}^{\min \, \, \{ j \, : \, r_{k-j}=1\}} \mathbbm{1}(s_{k-j}= s_k)\right)\]
to be the probability of generation of string $s$, when the urn is refreshed at every `1' in $r \in \{1,0\}^8$. Averaging over the distribution over refresh patterns, the probability of generation of string $s \in \{1,0\}^8$ is
\[g_R(s) = \sum_{r\in \{0,1\}^8} \left(p^{\| r\|_1} (1-p)^{8-\| r\|_1} \right) q(s|r).\]
 Finally, the probability that string $s$ was generated by a human is predicted to be
\[c_R(s) = \frac{g_R(s)}{g_R(s) + 1/256}.\]

The prediction rules based on \citet{gambler} have free parameters $\delta, \alpha \in  \mathbb{R}_+$. For Continuation, the probability that string $s \in \{1,0\}^7$ is followed by `1' is predicted to be
\[f_{RV}(s) = 0.5 - \alpha \sum_{t=0}^6 \delta^t (2s_{7-t}-1).\]
For Classification, define
\[g_{RV}(s) = 0.5 \cdot \prod_{k=2}^8 \left(0.5 - \alpha \sum_{t=0}^{k-2} \delta^t \mathbbm{1}\left(s_{k-t-1}=s_k \right) \right). \]
The probability that string $s \in \{1,0\}^8$ was generated by a human is predicted to be
\[c_{RV}(s) = \frac{g_{RV}(s)}{g_{RV}(s) + 1/256}.\]

\clearpage
\pagebreak

\section{Supplementary Material to Section \ref{sec:benchmark}}

\subsection{Different Cuts of the Data} \label{appDiffCut}

We repeat the main analysis in Section \ref{sec:main} using alternative cuts of the data. 

\emph{Only initial strings.} We consider a cut of the data in which we keep all subjects, but use only their first 25 strings. This selection accounts for potential fatigue in generation of the final strings, and leaves a total of 638 subjects and 15,950 strings. Prediction results for our main exercise are shown below using this alternative selection.

\begin{table}[H]
	\begin{minipage}{\columnwidth}
\begin{center}
\bigskip
\begin{tabular}{ccccc}	
\hline
&\multicolumn{2}{c}{Continuation}& \multicolumn{2}{c}{Classification}\\
	& Error & Completeness &  Error& Completeness \\
	\hline
	Bernoulli & 0.25 & 0 & 0.25 & 0 \vspace{2mm}\\
Rabin \& Vayanos (2010) & 0.2491 & 0.05 & 0.2480& 0.15 \\
&(0.0008)&&(0.0006)\vspace{2mm}\\
Table Lookup & 0.2326 &1 &    0.2367 & 1 \\
&(0.0030) & &(0.0030) \\
\hline
\end{tabular} 
\end{center}
		\centering
	\end{minipage}
\end{table}%

\emph{Chi-Squared Test.} For each subject, we conduct a Chi-squared test for the null hypothesis that their strings were generated under a Bernoulli process. We order subjects by $p$-values and remove the 100 subjects with the lowest $p$-values (subjects whose generated strings were most different from what we would expect under a Bernoulli process). This leaves a total of 538 subjects and 24,550 strings. Prediction results for our main exercise are shown below using this alternative selection.

\begin{table}[H]
	\begin{minipage}{\columnwidth}
\begin{center}
\bigskip
\begin{tabular}{ccccc}	
\hline
&\multicolumn{2}{c}{Continuation}& \multicolumn{2}{c}{Classification}\\
	& Error & Completeness &  Error& Completeness \\
	\hline
	Bernoulli & 0.25 & 0 & 0.25 & 0 \vspace{2mm}\\
Rabin \& Vayanos (2010) & 0.2491 & 0.12 & 0.2487& 0.15 \\
&(0.0005)&&(0.0005)\vspace{2mm}\\
Table Lookup & 0.2427 &1 &    0.2415  & 1 \\
&(0.0016) & &(0.0009) \\
\hline
\end{tabular} 
\end{center}
		\centering
	\end{minipage}
\end{table}%
\pagebreak
\clearpage

\subsection{Predict Conditioning on Different Numbers of Flips} \label{app:supplementary}
\vspace{30mm}
\begin{table}[H]
\caption{Continuation\textemdash Predict the $k$-th flip from the first $k-1$ realizations}
\label{tab:indexCont}
\bigskip
\begin{center}
\begin{tabular}{cccc}
\hline\hline
$k$ & Table Lookup &  \citet{gambler} & Completeness \\
\hline 
  2 & 0.2499    &  0.2500 & 0 \\
             &\footnotesize{(0.0004)} & \footnotesize{($<$0.0000)}  \\
  3 & 0.2489  &     0.2490  & 0.09\\
           &\footnotesize{(0.0005)} &  \footnotesize{(0.0004)}  \\
  4 & 0.2473  &    0.2482 & 0.67 \\
         &\footnotesize{(0.0012)} &   \footnotesize{(0.0011)} \\
  5 & 0.2433  &    0.2475 & 0.37\\
       &\footnotesize{(0.0031)} & \footnotesize{(0.0008)}   \\
  6 & 0.2461  &     0.2495 & 0.13\\
     &\footnotesize{(0.0024)} &  \footnotesize{(0.0004)} \\
  7 & 0.2424  &    0.2488 & 0.16 \\
   &\footnotesize{(0.0034)} & \footnotesize{(0.0006)}   \\
  8 & 0.2439  &    0.2492 & 0.13\\
&\footnotesize{(0.0019)} & \footnotesize{(0.0007)}\\
  \hline
\end{tabular}
\end{center}
\end{table}

\begin{table}
\caption{Classification\textemdash Separate Bernoulli strings of length $k$ from human-generated strings of length $k$}
\label{tab:indexClass}
\bigskip
\begin{center}
\begin{tabular}{cccc}\hline\hline
Length & Table Lookup & \citet{gambler} & Completeness\\
\hline 
  2 &   0.2494 &  0.2500 & 0\\
      &\footnotesize{(0.0001)} & \footnotesize{($<$0.0000)} \\
  3 & 0.2489  & 0.2500 & 0  \\
      &\footnotesize{(0.0001)} &  \footnotesize{($<$0.0000)} \\
  4 & 0.2484  & 0.2497 & 0.19 \\
      &\footnotesize{(0.0002)} & \footnotesize{(0.0003)}  \\
  5 & 0.2467  & 0.2491 & 0.27 \\
      &\footnotesize{(0.0002)} & \footnotesize{(0.0002)}\\
  6 & 0.2463  &  0.2492 & 0.22 \\
      &\footnotesize{(0.0004)} & \footnotesize{(0.0004)}\\
  7 &  0.2436  & 0.2489 & 0.18 \\
      &\footnotesize{(0.0007)} & \footnotesize{(0.0004)}\\
  8 &  0.2422  & 0.2488 & 0.15  \\
    &\footnotesize{(0.0010)} & \footnotesize{(0.0006)}\\
  \hline
\end{tabular}
\end{center}
\end{table}

\clearpage
\pagebreak

\subsection{Table Lookup Error} \label{app:bounds}
How close is the table lookup prediction error, estimated from $n$ samples, to the best possible prediction error?

\emph{Continuation.} For each string $s \in \{1,0\}^7$, let $p_s \in [0,1]$ be the true probability with which string $s$ is followed by `1.' Table lookup estimates this probability using the sample mean $\overline{p}^n_s$ (the frequency with which $s$ is followed by `1' in the training data). Notice that $\overline{p}^n_s \sim \text{Ber}(n_s, p_s)$, where $n_s$ is the number of times string $s$ is observed in the training data. The expected prediction error for string $s$ can be shown to be
\[\underbrace{\mathbb{E}_X[(X-p_s)^2]}_{\text{irreducible error}} + \mathbb{E}_{\mathcal{D}}[(p_s-\overline{p}^n_s)^2] \]
where $\mathcal{D}$ is the (random) training data, and $X \sim \text{Ber}(p_s)$ is the eighth flip. This expression is the sum of the irreducible error (which we want to estimate) and the variance of the table lookup estimator. Writing $q_s$ as the frequency with which string $s$ is generated as the initial substring, the expected table lookup prediction error is
\[\sum_{s\in \{1,0\}^7} q_s \cdot \left( \mathbb{E}_X[(X-p_s)^2] + \mathbb{E}_{\mathcal{D}}[(p_s-\overline{p}^n_s)^2]\right),\]
and the approximation error is
\[\sum_{s\in \{1,0\}^7} q_s \cdot \left( \mathbb{E}_{\mathcal{D}}[(p_s-\overline{p}^n_s)^2]\right) = \sum_{s\in \{1,0\}^7} q_s \cdot \frac{\overline{p}_s (1-\overline{p}_s)}{n_s}.\]
Substituting the empirical frequency $\hat{q}_s$ for the true frequency $q_s$ in the expression above, the approximation error for our main dataset is estimated to be 0.0014. Moreover, we can upper bound this quantity, using that $\sum_s q_s \cdot \frac{\overline{p}_s (1-\overline{p}_s)}{n_s} \leq \min_s \left(\frac{\overline{p}_s (1-\overline{p}_s)}{n_s} \right) = 0.0033$.

\emph{Classification.} For each string $s \in \{1,0\}^8$, let $p_s \in [0,1]$ be the true probability that string $s$ was generated by a human source, and let $\overline{p}^n_s$ be the proportion of instances of $s$ in the data that were generated by a human source. Using arguments similar to those above, the approximation error is 
\[\sum_{s\in \{1,0\}^8} q_s \cdot \left( \mathbb{E}_{\mathcal{D}}[(p_s-\overline{p}^n_s)^2]\right) = \sum_{s\in \{1,0\}^8} q_s \cdot \frac{\overline{p}_s (1-\overline{p}_s)}{n_s}.\]
noting that $\mathbb{E}(q_s)=n\cdot p_s$. Again substituting the empirical frequency $\hat{q}_s$ for the true frequency $q_s$ in the expression above, the approximation error for our main dataset is estimated to be 0.0031. An upper bound can be found using $  \sum_s q_s \cdot \frac{\overline{p}_s (1-\overline{p}_s)}{n_s} \leq \min_s \left(\frac{\overline{p}_s (1-\overline{p}_s)}{n_s} \right)=0.0066$.
\clearpage
\pagebreak 

\clearpage
\pagebreak
\clearpage
\pagebreak 

\section{Supplementary Material to Section \ref{sec:small} (Short Decision Tree Models)} \label{appendix:treesBehav}

\begin{figure}[H]
\begin{center}\title{Two-Split Decision Tree for Continuation\\(Trained On Behavioral Features)}
\includegraphics[scale=.6]{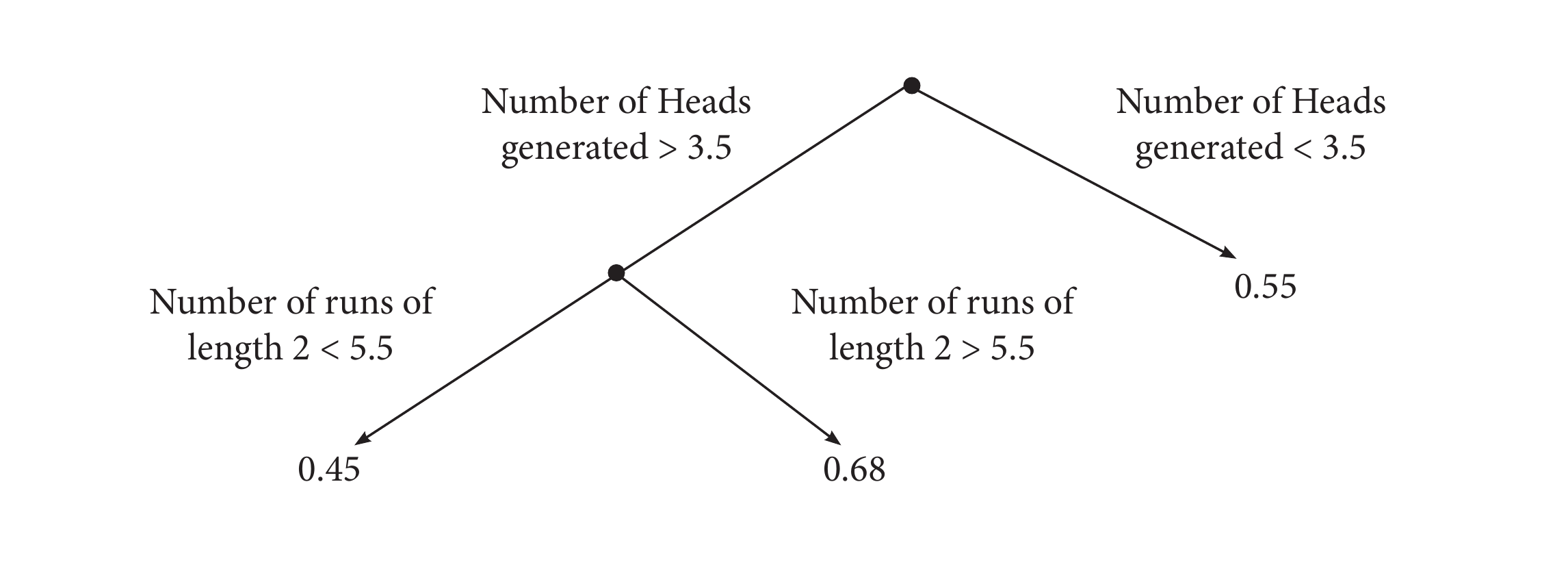}
\end{center}
\caption{\footnotesize{If the number of Heads generated in the first seven flips is less than 3.5, predict that the final flip is Heads with probability 0.55. Otherwise, look at the number of runs generated of length 2; if this is less than 5.5, predict 0.45; otherwise, predict 0.68.}}
\label{fig:BehavCont2}
\end{figure}

\vspace{20mm}
\begin{figure}[H]

\begin{center}
\title{Two-Split Decision Tree for Classification\\(Trained On Behavioral Features)}
\includegraphics[scale=.5]{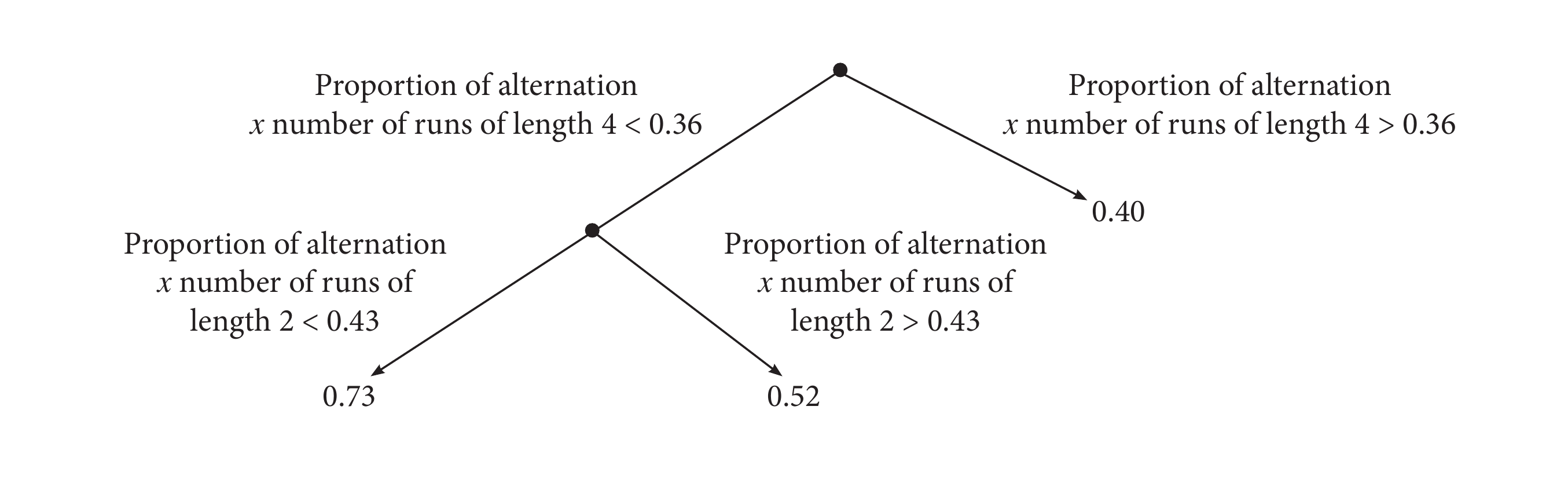}
\end{center}
\caption{\footnotesize{If the product of the proportion of alternation, and the number of runs of length four in the string, exceeds 0.36, then predict that the string is generated by a human subject with probability 0.40. Otherwise, predict 0.52 if the product of the proportion of alternation, and the number of runs of length two in the string, exceeds 0.43; if not, predict 0.73.}}
\label{fig:BehavClass2}
\end{figure}

\clearpage
\pagebreak

\begin{figure}
\begin{center}
\title{Two-Split Decision Tree for Continuation\\(Trained On Algorithmic Features)}
\includegraphics[scale=.5]{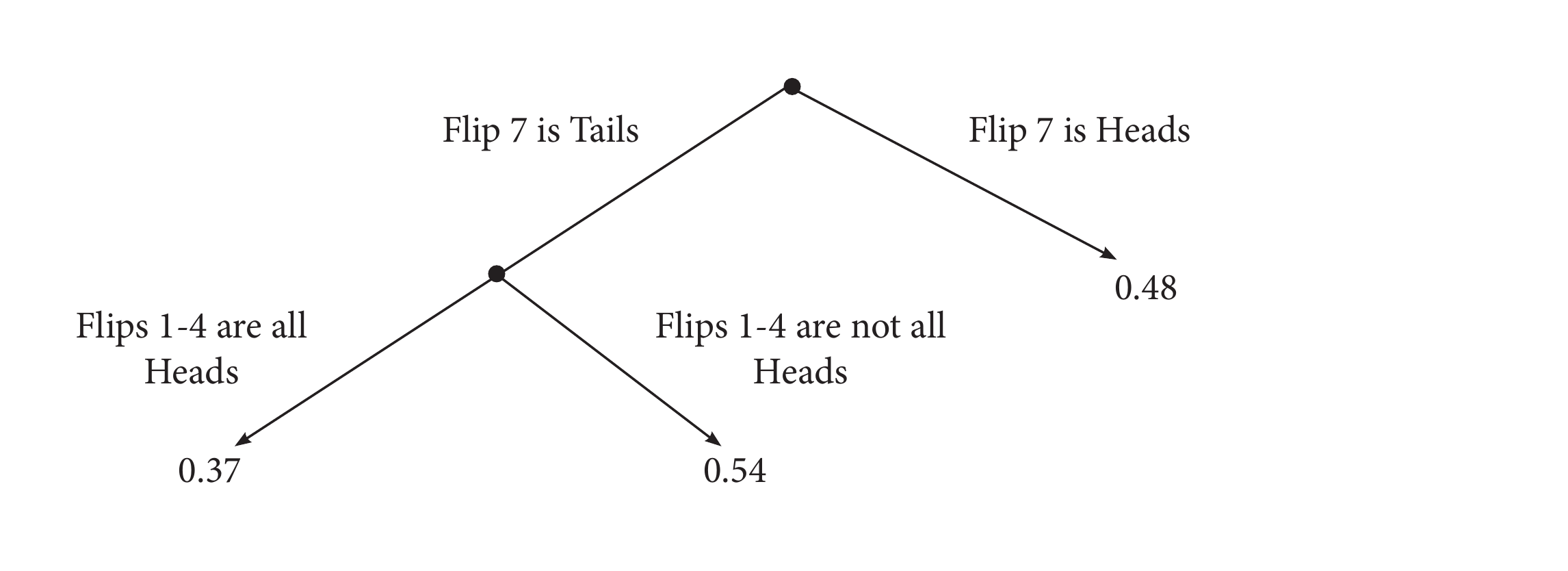}
\end{center}
\caption{\footnotesize{If the seventh flip is Heads, predict that the next flip is Heads with probability 0.48. Otherwise, if flips 1-4 are all Heads, predict 0.37, and if not, predict 0.54.}}
\label{fig:AlgCont2}
\end{figure}

\begin{figure}
\begin{center}
\title{Two-Split Decision Tree for Classification\\(Trained On Algorithmic Features)}
\includegraphics[scale=.5]{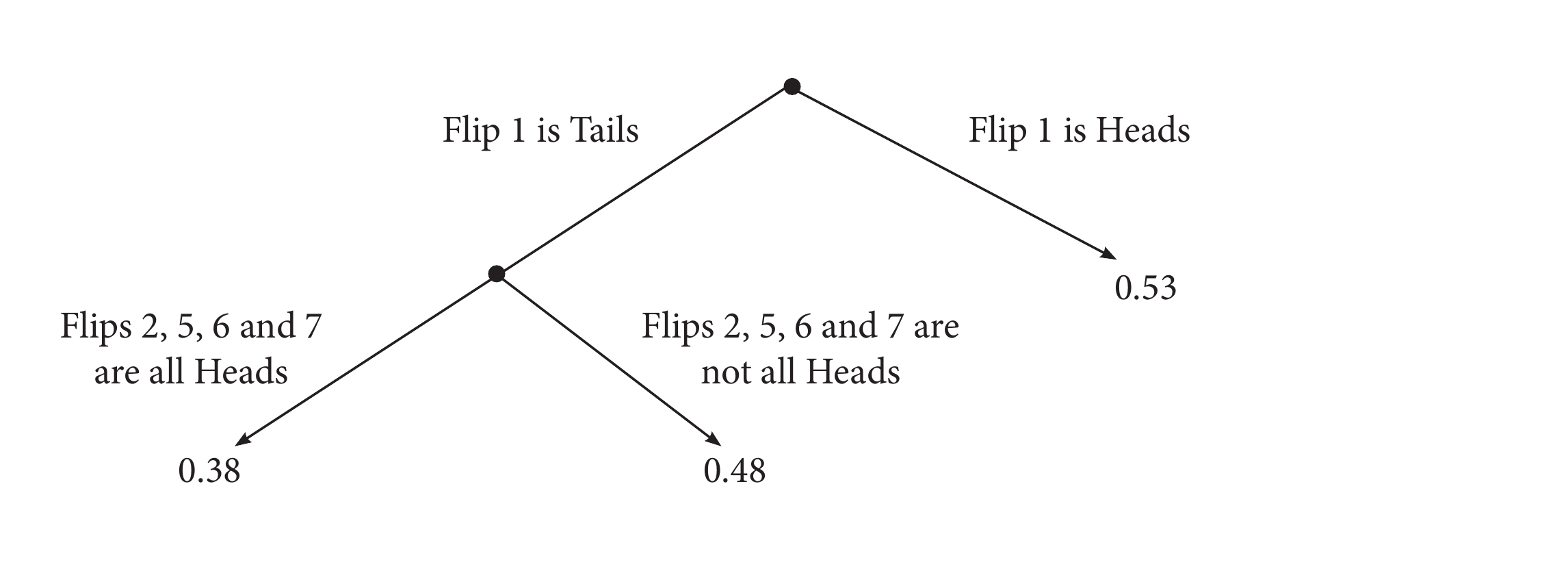}
\end{center}
\caption{\footnotesize{If the first flip is Heads, predict that the string was generated with a human with probability 0.53. Otherwise, if flips 2, 5, 6, and 7 are all Heads, predict 0.38, and if not, predict 0.48.}}
\label{fig:AlgClass2}
\end{figure}

\clearpage
\pagebreak

\section{Supplementary Material to Section \ref{sec:transfer} (Alternative Mappings)}

\begin{table}[H]%
\caption{Map `$H$' in the original coin flip data to `2' in $\{r,2\}$ and `$T$' in $\{H,T\}$. How do the Continuation prediction errors change from Table \ref{tab:Transfercont}?}
	\label{tab:Flippedcont}
	\bigskip
	\begin{minipage}{\columnwidth}
\begin{center}
\begin{tabular}{ccccc}	
\hline
&\multicolumn{2}{c}{$\{r,2\}^8$}& \multicolumn{2}{c}{$\{H,T\}^{15}$}\\
	& Error & Completeness &  Error& Completeness \\
	\hline
	Naive & 0.25 & 0 & 0.25 & 0  \\[2mm]
Rabin and Vayanos (2010) & 0.2496 & 0.06&   0.2480 & 0.15\\
&(0.0004)&&(0.0006) \\[2mm]
Table Lookup &0.2437 & 1 & 0.2370 & 1    \\
&(0.0011) &&(0.0024)  \\
\hline
\end{tabular} 
\end{center}
		\centering
	\end{minipage}
\end{table}%

\begin{table}[H]%
\caption{Map `$H$' in the original coin flip data to `2' in $\{r,2\}$ and `$T$' in $\{H,T\}$. How do the Classification prediction errors change from Table \ref{tab:Transferclass}?}
	\label{tab:Flippedclass}
	\bigskip
	\begin{minipage}{\columnwidth}
\begin{center}
\begin{tabular}{ccccc}	
\hline
&\multicolumn{2}{c}{$\{r,2\}^8$}& \multicolumn{2}{c}{$\{H,T\}^{15}$}\\
	& Error & Completeness &  Error& Completeness \\
	\hline
	Naive & 0.25 & 0 & 0.25 & 0  \\[2mm]
Rabin and Vayanos (2010) & 0.2494& 0.21 &  0.2488 & 0.12   \\
&(0.0007)&&(0.0006) \\[2mm]
Table Lookup &0.2471 & 1 &  0.2401 & 1    \\
&(0.0012) &&(0.0014)  \\
\hline
\end{tabular} 
\end{center}
		\centering
	\end{minipage}
\end{table}%

\begin{table}[H]%
\caption{Map `$H$' in the original coin flip data to `T'  in $\{H,T\}$ for the \citet{distributional} data. How do the prediction errors change from Tables \ref{tab:Transfercont} and \ref{tab:Transferclass}?}
	\label{tab:FlippedNB}
	\bigskip
	\begin{minipage}{\columnwidth}
\begin{center}
\begin{tabular}{ccccc}	
\hline
&\multicolumn{2}{c}{Continuation}& \multicolumn{2}{c}{Classification}\\
	& Error & Completeness &  Error& Completeness \\
	\hline
	Naive & 0.25 & 0 & 0.25 & 0  \\[2mm]
Rabin and Vayanos (2010) & 0.2491& 0.05 &    0.2497 & 0.03  \\
&(0.0004)&&(0.0004) \\[2mm]
Table Lookup &0.2336 & 1 &  0.2413 & 1    \\
&(0.0017) &&(0.0015)  \\
\hline
\end{tabular} 
\end{center}
		\centering
	\end{minipage}
\end{table}%

\clearpage
\pagebreak 

\section{Supplementary Material to Section \ref{sec:field} (Robustness Checks)} \label{app:field}

\subsection{Baseball Umpires}

In the main text, we predicted consecutive strings of umpire calls of length 6. Below we repeat the exercise, using consecutive strings of length 5 and length 4. There are 27,763 non-overlapping consecutive strings of length 5, and 56,312 non-overlapping consecutive strings of length 4.

\begin{table}[H]%
	\caption{Predicting sequences of umpire calls of length 5: \citet{gambler} explains 9-11\% of the explainable variation in the data.}
	\label{tab:umpire5}
	\bigskip
	\begin{minipage}{\columnwidth}
\begin{center}
\begin{tabular}{ccccc}	
\hline
&\multicolumn{2}{c}{Continuation}& \multicolumn{2}{c}{Classification}\\
	& Error & Completeness &  Error& Completeness \\
	\hline
	Naive &  0.2071\footnote{Standard error is 0.0027.} & 0 &0.25 &0 \vspace{2mm}\\
Rabin and Vayanos (2010) & 0.2066  & 0.11 &  0.2494 & 0.09 \\
&(0.0037)&&(0.0003)\vspace{2mm}\\
Table Lookup &   0.2027 & 1 &   0.2436 & 1\\
&(0.0027) &&(0.0011) \vspace{2mm}\\
\hline
\end{tabular} 
\end{center}
		\centering
	\end{minipage}
\end{table}%

\begin{table}[H]%
	\caption{Predicting sequences of umpire calls of length 4: \citet{gambler} explains 34-46\% of the explainable variation in the data.}
	\label{tab:umpire4}
	\bigskip
	\begin{minipage}{\columnwidth}
\begin{center}
\begin{tabular}{ccccc}	
\hline
&\multicolumn{2}{c}{Continuation}& \multicolumn{2}{c}{Classification}\\
	& Error & Completeness &  Error& Completeness \\
	\hline
	Naive &  0.2237\footnote{Standard error is 0.0027.} & 0 &0.25 &0 \vspace{2mm}\\
Rabin and Vayanos (2010) & 0.2209  & 0.34 &  0.2483  & 0.46  \\
&(0.0010)&&(0.0004)\vspace{2mm}\\
Table Lookup &   0.2154 & 1 &   0.2463 & 1\\
&(0.0029) &&(0.0005) \vspace{2mm}\\
\hline
\end{tabular} 
\end{center}
		\centering
	\end{minipage}
\end{table}%

\subsection{Rock-Paper-Scissors}

In the main text, we predicted consecutive strings of rock-paper-scissors throws of length 6. Below we repeat the exercise, using consecutive strings of length 5 and length 4. There are 61,335 non-overlapping consecutive strings of length 5, and 117,522 non-overlapping consecutive strings of length 4.

\begin{table}[H]%
	\caption{Predicting Rock-Paper-Scissors throws of length 5: \citet{gambler} explains 11-13\% of the explainable variation in the data.}
	\bigskip
	\begin{minipage}{\columnwidth}
\begin{center}
\begin{tabular}{ccccc}	
\hline
&\multicolumn{2}{c}{Continuation}& \multicolumn{2}{c}{Classification}\\
	& Error & Completeness &  Error& Completeness \\
	\hline
	Naive & 0.8165 & 0 & 0.25 & 0 \vspace{2mm}\\
Rabin and Vayanos (2010) &      0.8160   & 0.11 & 0.2492 & 0.13\\
&(0.0006)&&(0.0002)\vspace{2mm}\\
Table Lookup & 0.8120 & 1 &  0.2438 & 1\\
& (0.0006) &&(0.0008) \vspace{2mm}\\
\hline
\end{tabular} 
\end{center}
		\centering
	\end{minipage}
\end{table}%

\begin{table}[H]%
	\caption{Predicting Rock-Paper-Scissors throws of length 4: \citet{gambler} explains 15-18\% of the explainable variation in the data.}
	\bigskip
	\begin{minipage}{\columnwidth}
\begin{center}
\begin{tabular}{ccccc}	
\hline
&\multicolumn{2}{c}{Continuation}& \multicolumn{2}{c}{Classification}\\
	& Error & Completeness &  Error& Completeness \\
	\hline
	Naive & 0.8165 & 0 & 0.25 & 0 \vspace{2mm}\\
Rabin and Vayanos (2010) &      0.8159   & 0.15 & 0.2494 & 0.18\\
&(0.0002)&&(0.0002)\vspace{2mm}\\
Table Lookup & 0.8125 & 1 &  0.2466 & 1\\
& (0.0005) &&(0.0004) \vspace{2mm}\\
\hline
\end{tabular} 
\end{center}
		\centering
	\end{minipage}
\end{table}%
\clearpage 
\pagebreak

\clearpage
\pagebreak

\clearpage
\pagebreak


\clearpage

\begin{thebibliography}{11}
\newcommand{\enquote}[1]{``#1''}
\expandafter\ifx\csname natexlab\endcsname\relax\def\natexlab#1{#1}\fi

\bibitem[\protect\citeauthoryear{Bar-Hillel and Wagenaar}{Bar-Hillel and
  Wagenaar}{1991}]{barhillel}
\textsc{Bar-Hillel, M. and W.~Wagenaar} (1991): \enquote{The Perception of
  Randomness,} \emph{Advances in Applied Mathematics}.

\bibitem[\protect\citeauthoryear{Barberis, Shleifer, and Vishny}{Barberis
  et~al.}{1998}]{shleifer}
\textsc{Barberis, N., A.~Shleifer, and R.~Vishny} (1998): \enquote{A Model of
  Investor Sentiment,} \emph{Journal of Financial Economics}.

\bibitem[\protect\citeauthoryear{Batzilis, Jaffe, Levitt, List, and
  Picel}{Batzilis et~al.}{2016}]{RPS}
\textsc{Batzilis, D., S.~Jaffe, S.~Levitt, J.~A. List, and J.~Picel} (2016):
  \enquote{How Facebook Can Deepen our Understanding of Behavior in Strategic
  Settings: Evidence from a Million Rock-Paper-Scissors Games,} Working Paper.

\bibitem[\protect\citeauthoryear{Chen, Shue, and Moskowitz}{Chen
  et~al.}{2016}]{Shue}
\textsc{Chen, D., K.~Shue, and T.~Moskowitz} (2016): \enquote{Decision-Making
  under the Gambler's Fallacy: Evidence from Asylum Judges, Loan Officers, and
  Baseball Umpires,} \emph{Quarterly Journal of Economics}.

\bibitem[\protect\citeauthoryear{Hastie, Tibshirani, and Friedman}{Hastie
  et~al.}{2009}]{Hastie}
\textsc{Hastie, T., R.~Tibshirani, and J.~Friedman} (2009): \emph{The Elements
  of Statistical Learning}, Springer.

\bibitem[\protect\citeauthoryear{Nickerson and Butler}{Nickerson and
  Butler}{2009}]{distributional}
\textsc{Nickerson, R. and S.~Butler} (2009): \enquote{On Producing Random
  Sequences,} \emph{American Journal of Psychology}.

\bibitem[\protect\citeauthoryear{Peysakhovich and Naecker}{Peysakhovich and
  Naecker}{2017}]{Naecker}
\textsc{Peysakhovich, A. and J.~Naecker} (2017): \enquote{Using Methods from
  Machine Learning to Evaluate Behavioral Models of Choice Under Risk and
  Ambiguity,} \emph{Journal of Economic Behavior and Organization}.

\bibitem[\protect\citeauthoryear{Rabin}{Rabin}{2002}]{Rabin}
\textsc{Rabin, M.} (2002): \enquote{Inference by Believers in the Law of Small
  Numbers,} \emph{The Quarterly Journal of Economics}.

\bibitem[\protect\citeauthoryear{Rabin and Vayanos}{Rabin and
  Vayanos}{2010}]{gambler}
\textsc{Rabin, M. and D.~Vayanos} (2010): \enquote{The Gambler's and Hot-Hand
  Fallacies: Theory and Applications,} \emph{Review of Economic Studies}.

\bibitem[\protect\citeauthoryear{Rapaport and Budescu}{Rapaport and
  Budescu}{1997}]{budescu}
\textsc{Rapaport, A. and D.~Budescu} (1997): \enquote{Randomization in
  Individual Choice Behavior,} \emph{Psychological Review}.

\bibitem[\protect\citeauthoryear{Tversky and Kahneman}{Tversky and
  Kahneman}{1971}]{LSN}
\textsc{Tversky, A. and D.~Kahneman} (1971): \enquote{The Belief in the Law of
  Small Numbers,} \emph{Psychological Bulletin}.

\end{thebibliography}
\end{document}